\documentclass{article}

\usepackage[preprint]{neurips_2026}

\usepackage[dvipsnames]{xcolor}
\definecolor{linkColor}{RGB}{38,76,115}  % muted navy
\definecolor{citecolor}{RGB}{0,105,92}   % dark teal
\definecolor{urlColor}{RGB}{74,85,104}   % blue gray
\usepackage[colorlinks=true,linkcolor=linkColor,citecolor=citecolor,filecolor=linkColor,urlcolor=linkColor]{hyperref}

\usepackage[utf8]{inputenc} % allow utf-8 input
\usepackage[T1]{fontenc}    % use 8-bit T1 fonts
\usepackage{hyperref}       % hyperlinks
\usepackage{xurl}           % allow long reference URLs to wrap naturally
\usepackage{booktabs}       % professional-quality tables
\usepackage{amsfonts}       % blackboard math symbols
\usepackage{nicefrac}       % compact symbols for 1/2, etc.
\usepackage{microtype}      % microtypography
\usepackage{xcolor}         % colors

\usepackage{graphicx}
\usepackage{placeins}
\usepackage{tabularx}
\usepackage{multirow}
\usepackage{booktabs}
\usepackage{bbding}
\usepackage{caption}
\usepackage{pifont}
\usepackage{stfloats}
\usepackage{bm}
\usepackage{wrapfig}
\usepackage{colortbl}
\usepackage{enumitem}
\usepackage{indentfirst}
\usepackage{algorithm}
\usepackage{algpseudocode}
\usepackage{amsmath}
\usepackage{wrapfig}
\usepackage{booktabs}

\usepackage{fancyhdr} % for fancy headers and footers
\usepackage[most]{tcolorbox} % for colored boxes

\usepackage{caption}
\usepackage{subfigure}

\usepackage{amssymb}

\usepackage{xspace}

\usepackage{duckuments}

\usepackage{makecell}
\usepackage{fontawesome5}

\usepackage{adjustbox}

\usepackage{algorithm}
\usepackage{algpseudocode}

\title{
Chinese-Jev: Bringing System One Model to Chinese-Language Tasks
}

\vspace{-5pt}
\author{
  \vspace{-25pt}\\
  \textbf{
  Zexiao Wang$^{1,*}$,\quad
  Zihao Zhang$^{1,2,*}$,\quad
  Xudong Wang$^{2}$,\quad
  Pan Wang$^{3}$,\quad
  }\\[0pt]
  \textbf{
  Ziyi Ye$^{1}$, \quad
  Haoyu Zhao$^{1,2,\dagger}$,\quad
  Zuxuan Wu$^{1}$,\quad
  Shuicheng Yan$^{2}$
  }
  \vspace{3pt}\\
  $^1$Fudan University
  \quad
  $^2$National University of Singapore
  \quad \\
  $^3$University of Chinese Academy of Sciences
  \vspace{-4pt}
}

\begin{document}
\raggedbottom

\maketitle

\begingroup
\renewcommand{\thefootnote}{\fnsymbol{footnote}}

\footnotetext[1]{
Equal contribution.
}

\footnotetext[2]{
Corresponding author.
}

\endgroup

\begin{abstract}
System One models such as Jev offer an efficient alternative to generative language models for tasks that require decisions rather than open-ended responses. However, existing Jev models exhibit limited Chinese-language decision accuracy, restricting their utility in both general and specialized settings.
In this paper, we introduce \textsc{Chinese-Jev}, a System One model that addresses this gap through a unified data processing and training pipeline. 
Our data processing protocol converts heterogeneous Chinese-language annotations into probability targets over candidate options, enabling a shared training formulation across domains and question formats. 
To enable efficient inference, Chinese-Jev adopts a lightweight encoder-only backbone for text encoding and learns to score candidate answers through decision-oriented training.
To address the misalignment between the pre-training distribution and downstream Chinese-language scenarios, we first train the model on a general-purpose corpus of 10 million examples, then fine-tune it separately for the medical, legal, and financial domains.
To evaluate decision accuracy and calibration in both general and domain-specific Chinese-language settings, we introduce Chinese-Jev Bench (CJ-Bench).
After first-stage pre-training, Chinese-Jev exceeds the accuracy of the closed-source Jev model by 1.24\% on general-domain tasks while achieving a $20.3\times$ speedup. Subsequent domain-specific fine-tuning yields a 4.0\% accuracy improvement over Jev in medicine and retains 92\% of Jev's average accuracy across specialized domains, with a $17\times$ speedup and an average latency of only 15\, ms per example.
We further demonstrate on-device deployment of an INT8-quantized model on mobile devices, achieving an inference latency of approximately 1\, second per decision.
We will release the models, training data, benchmark, and data construction code.
The project is available at \href{https://gulucaptain.github.io/Chinese-Jev/}{https://gulucaptain.github.io/Chinese-Jev/}.
\end{abstract}

\section{Introduction}
\label{sec:intro}

An application does not always need a language model to write a response. It may need to route a request to a service~\citep{larson-etal-2019-evaluation,casanueva-etal-2020-efficient}, judge the relevance of a passage~\citep{nogueira2019reranking,khattab2020colbert}, or select an answer from a set of alternatives~\citep{lai-etal-2017-race,sun-etal-2020-investigating}. These tasks require language understanding, but their outputs are bounded. Large language models (LLMs) offer a flexible way to specify such tasks through instructions and examples~\citep{wei2022finetuned,sanh2022t0,pmlr-v202-longpre23a}. Repeatedly invoking a large model, however, can be costly when an application needs only a label or a score. The practical goal is to retain flexibility across tasks while reducing the cost of each decision.

Jev is a System One model designed for three forms of decision making: \texttt{choice} selects among candidate alternatives, \texttt{noul} determines the validity of a proposition, and \texttt{score} assigns an ordered rating~\citep{almeida2026jev}. Its structured outputs and associated probabilities can be used directly by software. Alongside the hosted Jev API, open implementations explore compact encoders and language-model-based decision systems~\citep{kotoba2026openjev,laya2026,cai2026openjev,palmer2026kev,lee2026semif}. Chinese-oriented releases include a model trained primarily on football-domain data~\citep{xuhaodev2026qwenjev} and a bilingual encoder for local agent decisions.

\begin{figure}[htb]
\centering
\includegraphics[width=\linewidth]{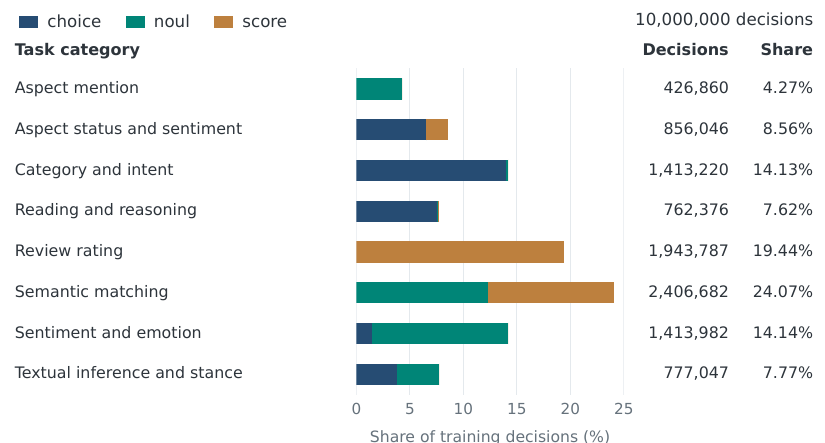}
\caption{General training corpus composition across eight task categories, ordered alphabetically. Stacked bars indicate category proportions, with colors denoting \texttt{choice}, \texttt{noul}, and \texttt{score}.}
\label{fig:data-coverage}
\end{figure}

Building on these efforts, we introduce \textsc{Chinese-Jev}, a compact System One model for Chinese-language decision-making across general tasks and the medical, legal, and financial domains. Central to this work is a unified supervision formulation that accommodates heterogeneous decision tasks. Existing Chinese benchmarks and instruction collections provide diverse source data~\citep{xu-etal-2020-clue,bai-etal-2025-coig}, but categorical labels, proposition judgments, and ordered ratings encode different forms of supervision. Our data processing protocol converts these annotations into probability distributions over candidate options while preserving label semantics and rating order. It also incorporates soft supervision when multiple ratings are available.
Applying this protocol yields a general corpus of 10 million decisions spanning eight task categories, together with dedicated corpora for medicine, law, and finance. Figure~\ref{fig:data-coverage} summarizes the composition of the general corpus.
To reduce leakage between training and evaluation, decisions derived from the same source material are assigned to the same data partition. This formulation supports option selection, proposition judgment, and ordinal rating through a shared candidate-scoring interface in a single forward pass.

Moreover, to address the mismatch between multilingual pre-training and downstream Chinese-language tasks while maintaining efficient inference, we adopt a two-stage training pipeline built on a lightweight encoder-only backbone~\citep{marone2025mmbert}. We first train \textsc{Chinese-Jev General} on a corpus of 10 million Chinese-language decisions spanning eight task categories, then independently fine-tune it on medical, legal, and financial corpora to obtain three domain specialists. All models retain the same architecture and decision interface, enabling domain specialization without increasing model size or changing how applications specify decisions.

We further introduce Chinese-Jev Bench (CJ-Bench), comprising 307,900 held-out decisions across general, medical, legal, and financial tasks, to evaluate decision accuracy, calibration, and inference latency under a common protocol. After first-stage pre-training, Chinese-Jev achieves 69.20\% accuracy on the general subset, exceeding the closed-source Jev model by 1.24\% in relative accuracy and reducing expected calibration error from 11.45\% to 3.78\%. Its measured latency is $20.3\times$ lower than that of the hosted Jev API. Following domain-specific fine-tuning, Chinese-Jev improves medical accuracy over Jev by 4.0\% and achieves 92\% of Jev's average accuracy across the three specialized domains, with an average latency of 15\, ms per decision and a $17\times$ reduction in measured latency.
Finally, an INT8-quantized deployment enables local inference in mobile devices at approximately 1\, second per decision, demonstrating the feasibility of on-device Chinese-language decision-making.

Our main contributions are:

\begin{itemize}[leftmargin=*,itemsep=3pt,topsep=4pt]
    \item \textbf{Chinese-Jev: a System One model for Chinese-language decision-making.}
    We develop Chinese-Jev through general Chinese-language pre-training followed by independent fine-tuning in medicine, law, and finance. The resulting general and specialist models share a unified decision interface that supports option selection, proposition judgment, and ordinal rating in a single forward pass.

    \item \textbf{A unified pipeline for Chinese-Jev training data construction.}
    We design a reusable pipeline that converts heterogeneous Chinese QA annotations into probability targets over candidate options for Chinese-Jev training. Using this pipeline, we construct a pre-training corpus of 10 million examples spanning eight task categories, together with downstream fine-tuning corpora.
    
    \item \textbf{CJ-Bench, empirical evaluation, and on-device deployment.}
    We introduce CJ-Bench, comprising 307,900 held-out decisions, to evaluate accuracy, calibration, and latency across general and specialized tasks. Chinese-Jev surpasses the closed-source Jev model in general-task accuracy and achieves 92\% of its average accuracy across specialized domains, with measured speedups of $20.3\times$ and $17\times$, respectively, relative to the hosted Jev API. An INT8 browser deployment further enables local smartphone inference at approximately 1\, second per decision.
\end{itemize}

\section{Related Work}
\label{sec:related}
\subsection{Generalist Text Classification}

Generalist text classification uses natural-language label descriptions to predict across tasks and label sets~\citep{yin-etal-2019-benchmarking,laurer2023universal,stepanov2025gliclass}. Input--label embedding methods learn compatibility between texts and labels to transfer to unseen categories~\citep{pappas-henderson-2019-gile}. Natural language inference (NLI) classifiers treat the input as a premise and each candidate label as a hypothesis~\citep{yin-etal-2019-benchmarking}. Universal NLI classifiers combine entailment data with classification datasets reformulated as premise--hypothesis pairs~\citep{laurer2023universal}. GLiClass jointly processes the input and all candidate labels in one encoder pass, allowing both text--label and label--label interactions~\citep{stepanov2025gliclass}. Chinese-Jev also encodes the input and candidate answers jointly, with separate readouts for choices, proposition judgments, and ordered ratings.

\subsection{Jev and Typed Decision Models}

Jev exposes \texttt{choice}, \texttt{noul}, and \texttt{score} decisions together with their associated probabilities~\citep{almeida2026jev}. Kotoba Labs' Open-Jev and Laya learn candidate-scoring functions over bidirectional encoder representations, with Laya's multilingual variant using mmBERT~\citep{kotoba2026openjev,laya2026}. Among language-model approaches, Zefan Cai's Open-Jev adapts a Qwen backbone with decision training~\citep{cai2026openjev}, Kev adds learned decision heads~\citep{palmer2026kev}, and Nimble fine-tunes predictions over allowed answer tokens~\citep{nimble2026}. SemIF reads probabilities from option-token logits~\citep{lee2026semif}, while AnyJev supports debiasing, post-hoc calibration, and lightweight fitted heads~\citep{anyjev2026}. Chinese-oriented releases include Qwen3-1.7B-Jev, which uses a language-model backbone and a learned decision head for football questions~\citep{xuhaodev2026qwenjev}, and MacJev, a bilingual encoder for tool routing and task-state checks~\citep{chaoliang2026macjev}. Chinese-Jev trains an encoder on general Chinese decisions and then adapts it separately to medicine, law, and finance.

\subsection{Multitask Supervision and Chinese Resources}

Multitask instruction tuning combines tasks through natural-language descriptions and shared training formats, enabling transfer beyond the tasks used for fine-tuning~\citep{wei2022finetuned,sanh2022t0,wang-etal-2022-super}. The Flan Collection examines task balancing, prompt diversity, and task reformulation, and shows the value of instruction-tuned checkpoints for subsequent task-specific adaptation~\citep{pmlr-v202-longpre23a}. For Chinese, CLUE covers general language understanding~\citep{xu-etal-2020-clue}, while C-Eval and CMMLU assess knowledge and reasoning across disciplines~\citep{NEURIPS2023_c6ec1844,li-etal-2024-cmmlu}. COIG-CQIA provides instruction-following data from real-world sources~\citep{bai-etal-2025-coig}. Specialized resources address biomedical language understanding, legal case retrieval, and financial knowledge and applications~\citep{zhang-etal-2022-cblue,li2024lecardv2,zhu-etal-2024-benchmarking}. Chinese-Jev converts selected source annotations into candidate-probability targets, preserving categorical labels, the order of grades, and distributions of human ratings.

\section{Methods}
\label{sec:method}

We develop \textsc{Chinese-Jev} to adapt compact System One decision models to broad Chinese-language tasks while retaining a unified interface across general and specialized domains. A unified data construction pipeline (Section~\ref{sec:data-pipeline}) maps diverse source annotations to probability distributions over candidate options while preserving their decision semantics. A general-to-domain training pipeline (Section~\ref{sec:model-training}) first adapts the model to broad Chinese supervision and then independently specializes it for medicine, law, and finance. Together, these components provide a common training and inference interface for \texttt{choice}, \texttt{noul}, and \texttt{score} decisions.

\subsection{Data Construction Pipeline}
\label{sec:data-pipeline}

\begin{figure}[!htbp]
\centering
\includegraphics[width=\linewidth]{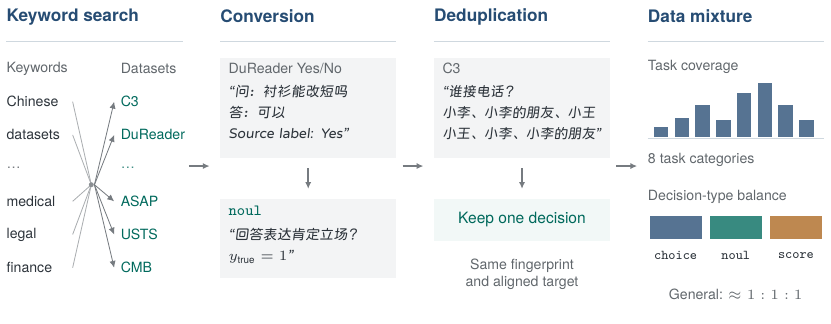}
\caption{Data construction pipeline comprising source discovery, annotation conversion, deduplication, and mixture selection. Examples illustrate mapping a ``DuReader Yes label'' to a \texttt{noul} stance target and deduplicating a C3 question across candidate permutations. The general training mixture spans eight task categories with approximately balanced budgets across the three decision types.}
\label{fig:data-pipeline}
\end{figure}

Figure~\ref{fig:data-pipeline} demonstrates our data construction pipeline, which consists of source discovery, annotation conversion, deduplication, and mixture construction. We first identify Chinese datasets whose annotations can be expressed as candidate-based decisions. Each source annotation is then converted into a common representation consisting of a context, an instruction, a decision type, a candidate set, and a target probability distribution. After removing duplicate and conflicting supervision, we construct the mixtures according to task coverage, available supervision, and decision-type budgets.

\paragraph{Source discovery and selection.}

We identify candidate datasets from published papers, author-maintained repositories, and Hugging Face using combinations of Chinese-language, task-specific, and domain-specific keywords. Aggregated collections are traced to their original sources, and we retain datasets with Chinese content, clear usage terms, and annotations that can be expressed as candidate-based decisions or ordered scores. Representative sources include C3 for reading comprehension~\cite{sun-etal-2020-investigating}, T2Ranking for relevance~\citep{xie2023t2ranking}, ASAP for review judgments~\citep{bu2021asap}, and USTS for similarity~\citep{wang2023collective}, alongside domain sources such as CMB~\citep{wang-etal-2024-cmb}, LeCaRDv2~\citep{li2024lecardv2}, and FinRE~\citep{li2019chinese}. We further construct 10,700 rule-based decisions for condition checking, counting, and grading, with construction details provided in Appendix~\ref{app:programmatic-data}.

\begin{figure}[htb]
\centering
\includegraphics[width=\linewidth]{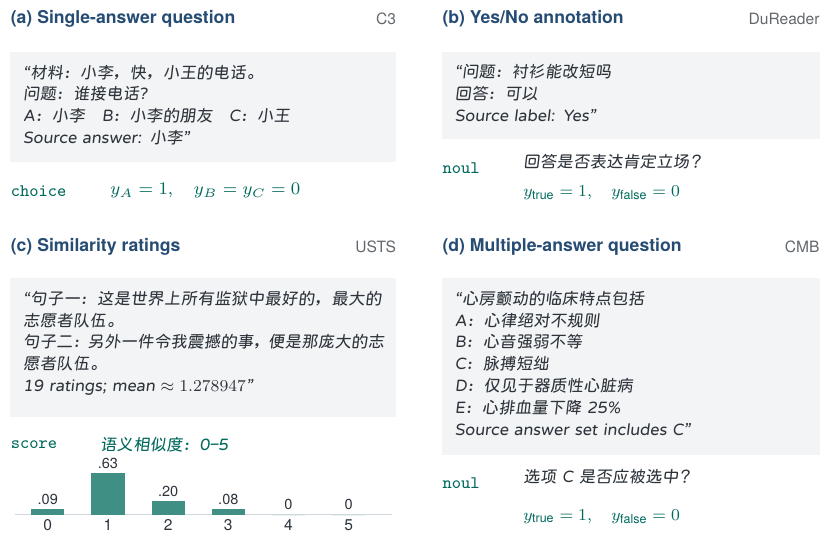}
\caption{Conversion of source annotations into decision targets: (a) single-answer questions to \texttt{choice}; (b) Yes/No annotations to \texttt{noul}; (c) ratings to \texttt{score}; and (d) multiple-answer questions to option-wise \texttt{noul}. DuReader labels answer stance. In (c), bars show the soft target obtained from 19 human ratings, rounded to two decimals. Examples are shortened for display.}
\label{fig:data-examples}
\end{figure}

\paragraph{Conversion to decision supervision.}
Each decision consists of a context $x$, an instruction $u$, a type $t$, a candidate list $C=(c_1,\ldots,c_K)$, and a target distribution $\mathbf{y}$ over $C$. Figure~\ref{fig:data-examples} illustrates four routes from source annotations to this format.

\emph{Single-answer questions and category labels $\rightarrow$ \texttt{choice}.}
For single-answer questions, we retain the original question and alternatives and assign probability one to the annotated answer. 
For categorical tasks, the candidate set comprises the source dataset's labels, with all target probability mass assigned to the annotated class.

\emph{True/false and proposition annotations $\rightarrow$ \texttt{noul}.}
We express the annotated judgment as an explicit proposition and map its binary label $z\in\{0,1\}$ to $\mathbf{y}=(1-z,z)$, ordered as \texttt{false} and \texttt{true}. For DuReader, the proposition asks whether a supplied answer expresses an affirmative stance: a Yes label maps to true without asserting the answer's factual correctness. Examples labeled Depends remain in the three-way \texttt{choice} task and are excluded from this \texttt{noul} conversion.

\emph{Ratings $\rightarrow$ \texttt{score}.}
For discrete ratings, we retain the original ordered scale and assign all target mass to the annotated level; relevance and quality grades follow the same rule. For continuous ratings, as in USTS, we distribute probability mass between adjacent integer levels for each rating and then average across raters. The soft target preserves the mean rating and inter-rater variation.

\emph{Multiple-answer questions $\rightarrow$ option-wise \texttt{noul}.}
We decompose each multiple-answer question into option-membership propositions. For a source question with $K$ options, let $A\subseteq\{1,\ldots,K\}$ be the correct option set and $z_j=\mathbf{1}[j\in A]$ indicate whether option $j$ belongs to it. The target is:
\begin{equation}
\mathbf{y}^{(j)}=(1-z_j,\,z_j),
\label{eq:multiselect-conversion}
\end{equation}
where the entries correspond to \texttt{false} and \texttt{true}. Each decision retains the full question and all alternatives but predicts membership for a single option. Complete multi-label annotations support the same conversion.

Inference annotations support relation selection or entailment judgments, while aspect annotations support mention detection and sentiment decisions. Synthetic rule tasks use the same formats, with targets computed from the underlying rules (Appendix~\ref{app:programmatic-data}).

\paragraph{Decision-level deduplication.}
For the general corpus, we define each decision by its task identifier $r$, type $t$, context $x$, instruction $u$, and candidates $C$, while excluding the target distribution so that conflicting supervision can be detected. After Unicode and whitespace normalization $N$, we canonicalize the candidates as:
\begin{equation}
\widetilde C_t=
\begin{cases}
\operatorname{sort}(N(C)), & t=\texttt{choice},\\
N(C), & t\in\{\texttt{noul},\texttt{score}\}.
\end{cases}
\label{eq:canonical-candidates}
\end{equation}
and compute the fingerprint:
\begin{equation}
h=H\!\left(r,t,N(x),N(u),\widetilde C_t\right).
\label{eq:decision-fingerprint}
\end{equation}
where $H$ denotes a hash function. Sorting makes \texttt{choice} fingerprints invariant to permutations of independent alternatives while preserving sensitivity to their content; the order of \texttt{score} levels and the false/true semantics of \texttt{noul} remain unchanged.
For matching fingerprints, we align \texttt{choice} targets by candidate text and retain a single decision when the targets agree, discarding the entire group when they conflict. Decisions that share source material, such as different questions derived from the same passage, may remain distinct but are assigned to the same data partition.
Appendix~\ref{app:deduplication} describes additional cases and checks against held-out material.

\paragraph{Task coverage and type balance.}
Decision-type budgets are defined separately for each corpus. For the general corpus, we allocate approximately one third of the training decisions to each of \texttt{choice}, \texttt{noul}, and \texttt{score}; within each type, source-level budgets preserve coverage of smaller tasks and diverse data sources before the remaining capacity is filled with eligible decisions. The legal and financial corpora follow the same balanced design, whereas the medical corpus contains a larger \texttt{noul} share due to its matching and multiple-label tasks; detailed domain statistics are reported in Appendix~\ref{app:domain-data}. We characterize the general corpus using eight task categories defined by prediction target and annotation semantics (Figure~\ref{fig:data-coverage}); for example, semantic matching includes similarity, equivalence, and retrieval relevance, while reading and reasoning cover contextual questions, logical reasoning, and explicit-rule decisions.

\paragraph{Data partitions and CJ-Bench.}
Each corpus has separate training, development, calibration, and test partitions, with all decisions derived from the same source material assigned to a single partition to reduce source-level leakage. Development data are used for model validation, while the calibration and test partitions remain disjoint from training and development. CJ-Bench is constructed exclusively from the held-out test partitions of the general, medical, legal, and financial corpora. Its General component covers all eight task categories with approximately equal decision-type budgets. Appendix~\ref{app:data-accounting} provides the complete partition statistics and benchmark composition.

\subsection{Chinese-Jev Training}
\label{sec:model-training}

\textsc{Chinese-Jev} adopts a lightweight encoder-only architecture for low-latency candidate scoring across general and domain-specific tasks. Given a decision $(x,u,t,C)$, a bidirectional mmBERT encoder~\citep{marone2025mmbert} jointly represents the input and candidate options, followed by a type-conditioned decision head that produces a probability distribution $\mathbf{p}$ over the candidates. We first train Chinese-Jev General on the general Chinese corpus and then independently specialize it for the medical, legal, and financial domains. The same architecture and training objective are retained throughout, preserving a consistent decision interface across all stages.

\begin{figure}[htb]
\centering
\includegraphics[width=\linewidth]{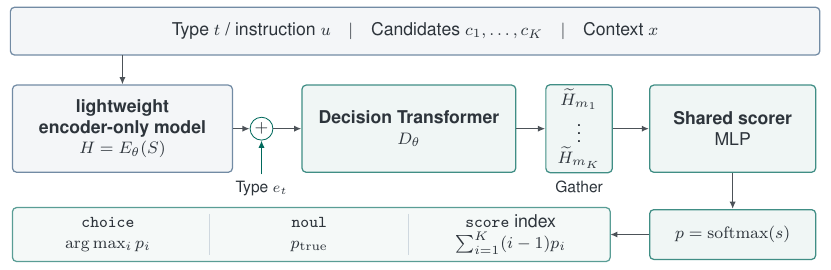}
\caption{Chinese-Jev architecture. A lightweight encoder-only model jointly processes the input and candidates, followed by a type-conditioned decision Transformer and a shared candidate scorer. All candidate probabilities are obtained in one forward pass. The \texttt{score} readout is an expected zero-based level index; source-scale ratings use the associated numerical level values.}
\label{fig:model-architecture}
\end{figure}

\paragraph{Model architecture.}
We serialize the decision type $t$, instruction $u$, all candidates $C$, and context $x$ into a single sequence $S$, placing a marker at position $m_i$ before each candidate $c_i$.

As shown in Figure~\ref{fig:model-architecture}, the encoder $E_\theta$ contextualizes the complete sequence:
\begin{equation}
H=E_\theta(S).
\label{eq:joint-encoding}
\end{equation}
We then add a learned type embedding $e_t$ at every position and apply the decision Transformer $D_\theta$:
\[
\widetilde{H}=D_\theta(H+\mathbf{1}e_t^\top).
\]
A shared MLP $f_\theta$ scores all candidate-marker representations in a single forward pass, with a softmax over the resulting logits yielding the decision probabilities:
\begin{equation}
s_i=f_\theta(\widetilde{H}_{m_i}),\qquad
p_i=\frac{\exp(s_i)}{\sum_{j=1}^{K}\exp(s_j)}.
\label{eq:candidate-scoring}
\end{equation}

\paragraph{Decision readouts.}
For \texttt{choice}, we return $\arg\max_i p_i$; for \texttt{noul}, we return $p_{\mathrm{true}}$. The \texttt{score} readout is the expected zero-based level index, $\sum_{i=1}^{K}(i-1)p_i$. On the source rating scale, the expectation is $\sum_i v_i p_i$, where $v_i$ is the numerical value of level $i$.

\paragraph{Training objective.}
We train the predicted distribution $\mathbf{p}$ against a target distribution $\mathbf{y}$ over the candidates defined in Section~\ref{sec:data-pipeline}. Following Laya's RLCD formulation~\citep{laya2026}, the objective combines supervised cross-entropy with a policy-gradient term over perturbed candidate logits. Both terms use $\mathbf{y}$. For a batch of $B$ decisions, where decision $b$ has $K_b$ candidates, the cross-entropy loss is:
\begin{equation}
\mathcal{L}_{\mathrm{CE}}=-\frac{1}{B}\sum_{b=1}^{B}\sum_{i=1}^{K_b}y_{bi}\log p_{bi}.
\label{eq:soft-ce}
\end{equation}
RLCD draws centered Gaussian perturbations of each decision's logits, reusing the same forward pass. Its reward uses log and spherical scores to measure agreement with the target distribution. For ordered \texttt{score} decisions, the reward also includes the ranked probability score (RPS) to account for distance along the ordered levels. The total objective is:
\begin{equation}
\mathcal{L}=\mathcal{L}_{\mathrm{CE}}+\mathcal{L}_{\mathrm{RL}}.
\label{eq:rlcd-loss}
\end{equation}
We provide the full RLCD perturbation, reward, and detached-advantage formulation in Appendix~\ref{app:rlcd}, and report the training configuration in Section~\ref{sec:experimental-setup}.

\section{Experiments} \label{sec:experiments}

\subsection{Experimental Setup}
\label{sec:experimental-setup}

\paragraph{Implementation details.}
We initialize the mmBERT encoder and decision head from Laya Multilingual~\citep{laya2026,laya2026multilingual}. The backbone is mmBERT-base with 22 layers and a hidden size of 768. The head contains two decision Transformer layers and a shared candidate scorer, bringing the model to approximately 322 million parameters.
Training uses full-parameter fine-tuning on eight NVIDIA H200 GPUs. We first train for one epoch on the general corpus to obtain Chinese-Jev General. Starting independently from this checkpoint, each specialist is trained for four epochs on its corresponding domain corpus. Medical specialization uses 3,000,000 decisions, while legal and financial specialization each use 24,000. The RLCD objective samples four logit perturbations per decision in both stages, shown in Appendix~\ref{app:rlcd}.

\paragraph{Benchmark and evaluation metrics.}
All comparisons use CJ-Bench (Section~\ref{sec:data-pipeline}), comprising General, Medical, Legal, and Finance components with 100,000, 200,000, 4,300, and 3,600 decisions, respectively. All models receive the same retained context and candidates and are evaluated on the same eligible examples within each component.
We mainly use three metrics for evaluation:
\textbf{1)} Accuracy (ACC) measures agreement between the highest-probability candidate and the gold label. \textbf{2)} Expected calibration error (ECE) measures the gap between this candidate's probability and empirical accuracy using 15 equal-width bins~\citep{guo2017calibration}. ACC and ECE apply to single-label examples, including hard \texttt{score} targets; soft targets are excluded. We retain each checkpoint's native probability transformations and apply no additional calibration on the target datasets.
\textbf{3)} Latency is the median end-to-end time per decision, measured on one NVIDIA H200 GPU at batch size one after warm-up. Timing includes input preparation and inference through the final candidate probabilities. For the hosted Jev API, we instead measure client-observed latency for one decision per request, including network round-trip time.

\paragraph{Baselines.}
Qwen3.5-2B~\citep{qwen2026qwen35two} is the general-purpose LLM baseline, evaluated on text inputs in non-thinking mode by scoring candidate labels. Open-Jev~\citep{cai2026openjev} denotes Zefan Cai's implementation, which adapts a Qwen backbone with decision training. SemIF~\citep{lee2026semif} reads decision probabilities from pretrained language models. Laya Multilingual~\citep{laya2026multilingual} uses a compact encoder. We access the official Jev model~\citep{almeida2026jev,typesafe2026models} through its hosted API. Each system supplies a candidate distribution for evaluation.

\subsection{Experimental Results}

\paragraph{Pre-training evaluation on general data.}
\label{sec:general-evaluation}

Table~\ref{tab:general-results} demonstrates the effectiveness of first-stage Chinese-language pre-training. Starting from Laya Multilingual, Chinese-Jev improves accuracy from 40.22\% to 69.20\%, a gain of 28.98 percentage points, while reducing ECE from 23.04\% to 3.78\%. These improvements are achieved without changing the model architecture or increasing measured inference latency: both models require a median of 14\,ms per decision. This comparison shows that Chinese-language training substantially improves decision accuracy and calibration without introducing additional inference cost.
Chinese-Jev achieves the highest accuracy and lowest ECE among the evaluated models, while matching the lowest measured latency. It exceeds the closed-source Jev model by 0.85 percentage points in accuracy, corresponding to a 1.24\% relative improvement, and reduces ECE by 67.0\%, from 11.45\% to 3.78\%. Under the same local timing protocol, Chinese-Jev is $20\times$ faster than Qwen3.5-2B while achieving substantially higher accuracy. Its measured latency is also $20.3\times$ lower than that of the hosted Jev API, although this comparison includes network round-trip time for Jev and therefore reflects end-to-end response latency rather than a direct comparison of model inference speed.

% \begin{table}[htbp]
% \centering
% \caption{Results on the general subset of the Chinese-Jev Benchmark after first-stage pre-training. The best and second-best results are highlighted in \textbf{bold} and \underline{underlined}, respectively.}
% \label{tab:general-results}
% \small
% \renewcommand{\arraystretch}{1.32}
% \setlength{\tabcolsep}{6pt}
% \begin{tabularx}{\linewidth}{>{\raggedright\arraybackslash}p{0.42\linewidth}*{3}{>{\centering\arraybackslash}X}}
% \toprule
% \rowcolor{linkColor!6}
% \textbf{Model} & \makecell{\textbf{ACC} $\uparrow$\\\footnotesize (\%)} & \makecell{\textbf{ECE} $\downarrow$\\\footnotesize (\%)} & \makecell{\textbf{Latency} $\downarrow$\\\footnotesize (ms)} \\
% \midrule
% Qwen3.5-2B & 32.95&43.68 & 280 \\
% Open-Jev & 52.33 &\underline{10.54} & 60 \\
% SemIF &44.21 &27.27 & 57 \\
% Laya Multilingual & 40.22 &23.04 & \underline{14} \\
% Jev\textsuperscript{$\dagger$} & \underline{68.35} & 11.45& 284\\
% \midrule
% \rowcolor{citecolor!7}
% \textbf{Chinese-Jev General} & \textbf{69.20}& \textbf{3.78} & \textbf{14}\\
% \bottomrule
% \end{tabularx}
% \par\vspace{4pt}
% \begin{minipage}{\linewidth}
% \footnotesize
% $\dagger$ Jev uses a hosted API; its latency includes network round-trip time.
% \end{minipage}
% \end{table}

\begin{table}[t]
\centering
\caption{Results on the general subset of Chinese-Jev Benchmark.
Chinese-Jev General is evaluated after first-stage pre-training.
The best and second-best results are shown in \textbf{bold} and
\underline{underlined}, respectively; ties share the same rank.
$\dagger$ Jev uses a hosted API, and its latency includes
network round-trip time.}
\label{tab:general-results}
\small
\renewcommand{\arraystretch}{1.15}
\setlength{\tabcolsep}{0pt}
\begin{tabular*}{\linewidth}{
    @{\extracolsep{\fill}} l c c c @{}
}
\toprule
\textbf{Model}
& \textbf{ACC (\%) $\uparrow$}
& \textbf{ECE (\%) $\downarrow$}
& \textbf{Latency (ms) $\downarrow$} \\
\midrule
Qwen3.5-2B~\citep{qwen2026qwen35two}
& 32.95 & 43.68 & 280 \\
Open-Jev~\citep{cai2026openjev}
& 52.33 & \underline{10.54} & 60 \\
SemIF~\citep{lee2026semif}
& 44.21 & 27.27 & \underline{57} \\
Laya Multilingual~\citep{laya2026multilingual}
& 40.22 & 23.04 & \textbf{14} \\
Jev\textsuperscript{$\dagger$}~\citep{almeida2026jev}
& \underline{68.35} & 11.45 & 284 \\
\midrule
% \rowcolor{citecolor!5}
\textbf{Chinese-Jev General}
& \textbf{69.20}\,{\scriptsize (+0.85)}
& \textbf{3.78}\,{\scriptsize (-6.76)}
& \textbf{14}\, \\
\bottomrule
\end{tabular*}
\end{table}

\paragraph{Domain-specific evaluation.}
\label{sec:domain-evaluation}

Table~\ref{tab:domain-results} shows that domain-specific fine-tuning substantially improves decision accuracy beyond general Chinese-language pre-training. Relative to Chinese-Jev General, the specialists improve accuracy by 47.88, 21.34, and 29.33 percentage points on Medical, Legal, and Finance, respectively, raising average accuracy from 37.97\% to 70.82\%. General pre-training alone provides uneven benefits across domains: on Finance, Chinese-Jev General achieves 35.39\% accuracy, below the multilingual initialization's 38.31\%. These results highlight the value of supervision tailored to the target domain beyond broad Chinese-language training.

Chinese-Jev Specialist outperforms all four open baselines in accuracy across the three domains. On Medical, it achieves 87.65\% accuracy, exceeding the closed-source Jev model by 3.36 percentage points, or 4.0\% in relative terms. The Legal and Finance specialists reach 60.10\% and 64.72\%, respectively, after fine-tuning on 24,000 decisions per domain, although both remain below Jev. Across domains, the specialists achieve 92.0\% of Jev's average accuracy while retaining a median latency of 15\,ms per decision in each domain. This corresponds to $17.5$--$29.2\times$ speedups over Qwen3.5-2B under the same local timing protocol. Their average latency is also approximately one-seventeenth that of the hosted Jev API, whose measurements include network round-trip time.
The calibration benefits of specialization are less consistent. ECE decreases from 17.33\% to 7.25\% on Medical and from 14.43\% to 11.55\% on Finance, but increases from 15.37\% to 17.45\% on Legal despite the accuracy gain. Across models, the Medical specialist achieves the highest accuracy, whereas Open-Jev obtains the lowest ECE. Moreover, the specialists' average ECE remains higher than Jev's (12.08\% versus 5.57\%).
These results highlight the need to assess calibration separately from accuracy after domain adaptation.

\begin{table*}[t]
\centering
\caption{Domain-specific results on CJ-Bench.
Chinese-Jev-G and Chinese-Jev-S denote the general model and
the corresponding domain specialist, respectively.
ACC and ECE are reported in \%, and latency (Lat.) in ms.
Avg.\ denotes the unweighted mean of domain-level metrics.
The best and second-best results are shown in \textbf{bold}
and \underline{underlined}, respectively.
$\dagger$ Jev's hosted API latency includes network round-trip time.}
\label{tab:domain-results}

\small
\renewcommand{\arraystretch}{1.12}
\setlength{\tabcolsep}{3pt}

\begin{adjustbox}{max width=\linewidth}
\begin{tabular}{@{}l*{12}{r}@{}}
\toprule
& \multicolumn{3}{c}{\textbf{Medical}}
& \multicolumn{3}{c}{\textbf{Legal}}
& \multicolumn{3}{c}{\textbf{Finance}}
& \multicolumn{3}{c}{\textbf{Average}} \\
\cmidrule(lr){2-4}
\cmidrule(lr){5-7}
\cmidrule(lr){8-10}
\cmidrule(l){11-13}
Model
& ACC $\uparrow$ & ECE $\downarrow$ & Lat. $\downarrow$
& ACC $\uparrow$ & ECE $\downarrow$ & Lat. $\downarrow$
& ACC $\uparrow$ & ECE $\downarrow$ & Lat. $\downarrow$
& ACC $\uparrow$ & ECE $\downarrow$ & Lat. $\downarrow$ \\
\midrule
Qwen3.5-2B~\citep{qwen2026qwen35two}
& 53.88 & 28.71 & 263
& 31.53 & 44.82 & 438
& 38.11 & 49.12 & 266
& 41.17 & 40.88 & 322.33 \\

Open-Jev~\citep{cai2026openjev}
& 73.08 & \textbf{2.88} & 53
& 44.07 & 23.15 & 110
& 48.11 & \underline{10.19} & 73
& 55.09 & \underline{12.07} & 78.67 \\

SemIF~\citep{lee2026semif}
& 62.17 & 10.39 & \underline{51}
& 43.07 & 19.99 & \underline{49}
& 47.89 & 18.45 & \underline{54}
& 51.04 & 16.28 & \underline{51.33} \\

Laya~\citep{laya2026multilingual}
& 24.55 & 43.05 & \textbf{15}
& 34.72 & 31.49 & \textbf{15}
& 38.31 & 26.18 & \textbf{15}
& 32.53 & 33.57 & \textbf{15.00} \\

Jev\textsuperscript{$\dagger$}~\citep{almeida2026jev}
& \underline{84.29} & \underline{5.70} & 267
& \textbf{68.35} & \textbf{8.03} & 256
& \textbf{78.39} & \textbf{2.98} & 246
& \textbf{77.01} & \textbf{5.57} & 256.33 \\
\midrule
Chinese-Jev-G
& 39.77 & 17.33 & \textbf{15}
& 38.76 & \underline{15.37} & \textbf{15}
& 35.39 & 14.43 & \textbf{15}
& 37.97 & 15.71 & \textbf{15.00} \\

\rowcolor{citecolor!5}
\textbf{Chinese-Jev-S}
& \textbf{87.65} & 7.25 & \textbf{15}
& \underline{60.10} & 17.45 & \textbf{15}
& \underline{64.72} & 11.55 & \textbf{15}
& \underline{70.82} & 12.08 & \textbf{15.00} \\
\bottomrule
\end{tabular}
\end{adjustbox}
\end{table*}

\subsection{Application: Interactive Decisions on Mobile Devices}
\label{sec:mobile-application}

\begin{figure}[!htbp]
\centering
\includegraphics[width=0.9\linewidth]{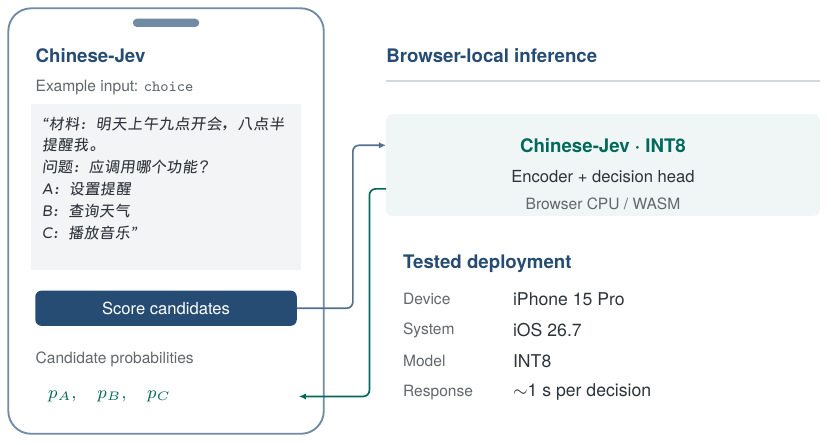}
\caption{On-device Chinese-Jev browser prototype. A Chinese reminder request and editable candidates are scored locally by the INT8 model. The interface is schematic, with $p_A$, $p_B$, and $p_C$ standing for returned probabilities. The tested iPhone 15 Pro configuration responds in approximately one second per decision.}
\label{fig:mobile-application}
\end{figure}

Our mobile browser prototype enables users to submit Chinese questions with context and editable candidate answers, and obtain option probabilities directly on their smartphones.
Figure~\ref{fig:mobile-application} illustrates this workflow using a meeting message with three candidate actions: setting a reminder, checking the weather, and playing music.
On an iPhone 15 Pro running iOS 26.7, the INT8-quantized model achieves an inference latency of approximately 1\,s per decision. Both tokenization and inference execute locally in the browser, with downloaded model files cached for reuse. Once the model is loaded, subsequent queries are processed entirely on-device without transmitting input text to a remote inference service.

\section{Conclusion}
\label{sec:conclusion}

We introduced \textsc{Chinese-Jev}, a System One model for Chinese-language decision-making across general tasks and specialized domains. Our unified data construction pipeline supports large-scale Chinese pre-training and subsequent fine-tuning in medicine, law, and finance. We also introduced CJ-Bench to evaluate decision accuracy, calibration, and inference latency. Chinese-Jev surpasses the closed-source Jev model on general tasks and in medicine, while achieving 92\% of its average accuracy across specialized domains. Measured latency improves by approximately $20.3\times$ on general tasks and $17\times$ across specialized domains relative to the hosted Jev API, whose latency includes network overhead. An INT8 mobile browser deployment further demonstrates on-device inference at approximately 1\, second per decision. Remaining accuracy gaps in law and finance, together with uneven calibration gains after specialization, motivate further work on domain adaptation and confidence calibration. We will release the models, training data, benchmark, and data construction code to support reproducible research.

\bibliographystyle{chinesejev}
\begingroup
\setlength{\bibsep}{4pt plus 1pt minus 1pt}
\bibliography{neurips_2026}

@inproceedings{casanueva-etal-2020-efficient,
  title     = {Efficient Intent Detection with Dual Sentence Encoders},
  author    = {Casanueva, I{\~n}igo and Tem{\v{c}}inas, Tadas and Gerz, Daniela and Henderson, Matthew and Vuli{\'c}, Ivan},
  booktitle = {Proceedings of the 2nd Workshop on Natural Language Processing for Conversational AI},
  year      = {2020},
  publisher = {Association for Computational Linguistics},
  pages     = {38--45},
  doi       = {10.18653/v1/2020.nlp4convai-1.5},
  url       = {https://aclanthology.org/2020.nlp4convai-1.5/}
}

@misc{nogueira2019reranking,
  title         = {Passage Re-ranking with {BERT}},
  author        = {Nogueira, Rodrigo and Cho, Kyunghyun},
  year          = {2019},
  eprint        = {1901.04085},
  archivePrefix = {arXiv},
  primaryClass  = {cs.IR},
  howpublished  = {\url{https://arxiv.org/abs/1901.04085}}
}

@misc{typesafe2026models,
  author       = {{TypeSafe}},
  title        = {{Jev}: Models and Language Support},
  year         = {2026},
  howpublished = {\url{https://docs.typesafe.ai/models}},
  note         = {Official documentation, accessed September 27, 2026}
}

@misc{xuhaodev2026qwenjev,
  author       = {{xuhaodev}},
  title        = {{Qwen3-1.7B-Jev}},
  year         = {2026},
  howpublished = {\url{https://huggingface.co/xuhaodev/Qwen3-1.7B-Jev}},
  note         = {Model card, revision \href{https://huggingface.co/xuhaodev/Qwen3-1.7B-Jev/blob/22e7aa20a4aaeda3a08999c163193c02c9607c9b/README.md}{\texttt{22e7aa2}}}
}

@misc{chaoliang2026macjev,
  author       = {{chaoliangUNSW}},
  title        = {{MacJev-322M-4K-Laya}},
  year         = {2026},
  howpublished = {\url{https://huggingface.co/chaoliangUNSW/MacJev-322M-4K-Laya}},
  note         = {Model card, revision \href{https://huggingface.co/chaoliangUNSW/MacJev-322M-4K-Laya/blob/92b182e653ebd8facd04ec9bddf6f63c4a7b7c22/README.md}{\texttt{92b182e}}}
}

@misc{laya2026,
  author       = {{NandhaKishorM}},
  title        = {Laya},
  year         = {2026},
  howpublished = {\url{https://github.com/NandhaKishorM/laya}},
  note         = {Version 0.3.11, commit \href{https://github.com/NandhaKishorM/laya/tree/1e28ac20c0896b1c37a744cd11f740eb98f8b178}{\texttt{1e28ac2}}}
}

@inproceedings{bu2021asap,
  title     = {{ASAP}: A {Chinese} Review Dataset Towards Aspect Category Sentiment Analysis and Rating Prediction},
  author    = {Bu, Jiahao and Ren, Lei and Zheng, Shuang and Yang, Yang and Wang, Jingang and Zhang, Fuzheng and Wu, Wei},
  booktitle = {Proceedings of the 2021 Conference of the North American Chapter of the Association for Computational Linguistics: Human Language Technologies},
  year      = {2021},
  publisher = {Association for Computational Linguistics},
  pages     = {2069--2079},
  doi       = {10.18653/v1/2021.naacl-main.167},
  url       = {https://aclanthology.org/2021.naacl-main.167/}
}

@misc{xie2023t2ranking,
  title         = {{T2Ranking}: A Large-scale {Chinese} Benchmark for Passage Ranking},
  author        = {Xie, Xiaohui and Dong, Qian and Wang, Bingning and Lv, Feiyang and Yao, Ting and Gan, Weinan and Wu, Zhijing and Li, Xiangsheng and Li, Haitao and Liu, Yiqun and Ma, Jin},
  year          = {2023},
  eprint        = {2304.03679},
  archivePrefix = {arXiv},
  primaryClass  = {cs.IR},
  howpublished  = {\url{https://arxiv.org/abs/2304.03679}}
}

@article{wang2023collective,
  title     = {Collective Human Opinions in Semantic Textual Similarity},
  author    = {Wang, Yuxia and Tao, Shimin and Xie, Ning and Yang, Hao and Baldwin, Timothy and Verspoor, Karin},
  journal   = {Transactions of the Association for Computational Linguistics},
  volume    = {11},
  year      = {2023},
  pages     = {997--1013},
  doi       = {10.1162/tacl_a_00584},
  url       = {https://aclanthology.org/2023.tacl-1.56/}
}

@article{sun-etal-2020-investigating,
    title = "Investigating Prior Knowledge for Challenging {C}hinese Machine Reading Comprehension",
    author = "Sun, Kai  and
      Yu, Dian  and
      Yu, Dong  and
      Cardie, Claire",
    editor = "Johnson, Mark  and
      Roark, Brian  and
      Nenkova, Ani",
    journal = "Transactions of the Association for Computational Linguistics",
    volume = "8",
    year = "2020",
    address = "Cambridge, MA",
    publisher = "MIT Press",
    url = "https://aclanthology.org/2020.tacl-1.10/",
    doi = "10.1162/tacl_a_00305",
    pages = "141--155"
}

@inproceedings{wang-etal-2024-cmb,
    title = "{CMB}: A Comprehensive Medical Benchmark in {C}hinese",
    author = "Wang, Xidong  and
      Chen, Guiming Hardy  and
      Song, Dingjie  and
      Zhang, Zhiyi  and
      Chen, Zhihong  and
      Xiao, Qingying  and
      Jiang, Feng  and
      Li, Jianquan  and
      Wan, Xiang  and
      Wang, Benyou  and
      Li, Haizhou",
    editor = "Duh, Kevin  and
      Gomez, Helena  and
      Bethard, Steven",
    booktitle = "Proceedings of the 2024 Conference of the North American Chapter of the Association for Computational Linguistics: Human Language Technologies (Volume 1: Long Papers)",
    month = jun,
    year = "2024",
    address = "Mexico City, Mexico",
    publisher = "Association for Computational Linguistics",
    url = "https://aclanthology.org/2024.naacl-long.343/",
    doi = "10.18653/v1/2024.naacl-long.343",
    pages = "6184--6205"
}

@inproceedings{zhang-etal-2022-cblue,
    title = "{CBLUE}: A {C}hinese Biomedical Language Understanding Evaluation Benchmark",
    author = "Zhang, Ningyu  and
      Chen, Mosha  and
      Bi, Zhen  and
      Liang, Xiaozhuan  and
      Li, Lei  and
      Shang, Xin  and
      Yin, Kangping  and
      Tan, Chuanqi  and
      Xu, Jian  and
      Huang, Fei  and
      Si, Luo  and
      Ni, Yuan  and
      Xie, Guotong  and
      Sui, Zhifang  and
      Chang, Baobao  and
      Zong, Hui  and
      Yuan, Zheng  and
      Li, Linfeng  and
      Yan, Jun  and
      Zan, Hongying  and
      Zhang, Kunli  and
      Tang, Buzhou  and
      Chen, Qingcai",
    editor = "Muresan, Smaranda  and
      Nakov, Preslav  and
      Villavicencio, Aline",
    booktitle = "Proceedings of the 60th Annual Meeting of the Association for Computational Linguistics (Volume 1: Long Papers)",
    month = may,
    year = "2022",
    address = "Dublin, Ireland",
    publisher = "Association for Computational Linguistics",
    url = "https://aclanthology.org/2022.acl-long.544/",
    doi = "10.18653/v1/2022.acl-long.544",
    pages = "7888--7915"
}

@inproceedings{NEURIPS2023_a48ad12d,
 author = {Liu, Junling and Zhou, Peilin and Hua, Yining and Chong, Dading and Tian, Zhongyu and Liu, Andrew and Wang, Helin and You, Chenyu and Guo, Zhenhua and ZHU, LEI and Li, Michael Lingzhi},
 booktitle = {Advances in Neural Information Processing Systems},
 doi = {10.52202/075280-2283},
 editor = {A. Oh and T. Naumann and A. Globerson and K. Saenko and M. Hardt and S. Levine},
 pages = {52430--52452},
 publisher = {Curran Associates, Inc.},
 title = {Benchmarking Large Language Models on CMExam - A comprehensive Chinese Medical Exam Dataset},
 url = {https://proceedings.neurips.cc/paper_files/paper/2023/file/a48ad12d588c597f4725a8b84af647b5-Paper-Datasets_and_Benchmarks.pdf},
 volume = {36},
 year = {2023}
}

@inproceedings{guo2017calibration,
  title = {On Calibration of Modern Neural Networks},
  author = {Guo, Chuan and Pleiss, Geoff and Sun, Yu and Weinberger, Kilian Q.},
  booktitle = {Proceedings of the 34th International Conference on Machine Learning},
  series = {Proceedings of Machine Learning Research},
  volume = {70},
  pages = {1321--1330},
  year = {2017},
  publisher = {PMLR},
  url = {https://proceedings.mlr.press/v70/guo17a.html}
}

@misc{qwen2026qwen35two,
  author = {{Qwen Team}},
  title = {{Qwen3.5-2B}},
  year = {2026},
  howpublished = {\url{https://huggingface.co/Qwen/Qwen3.5-2B}},
  note = {Model card, accessed September 28, 2026}
}

@misc{laya2026multilingual,
  author = {{Convai Innovations}},
  title = {{Laya Multilingual}},
  year = {2026},
  howpublished = {\href{https://huggingface.co/convaiinnovations/laya-multilingual/blob/e4e9ddf21a7b1903b7acffd8814ad4307bf63a67/README.md}{Hugging Face model card (revision \texttt{e4e9ddf21a7b})}},
  note = {Accessed September 27, 2026}
}

@misc{marone2025mmbert,
  title = {{mmBERT: A Modern Multilingual Encoder with Annealed Language Learning}},
  author = {Marone, Marc and Weller, Orion and Fleshman, William and Yang, Eugene and Lawrie, Dawn and Van Durme, Benjamin},
  year = {2025},
  eprint = {2509.06888},
  archivePrefix = {arXiv},
  howpublished = {\url{https://arxiv.org/abs/2509.06888v1}},
  note = {Preprint, v1}
}

@misc{almeida2026jev,
  author = {Almeida, Diogo},
  title = {{Introducing System One Models \& Jev}},
  year = {2026},
  howpublished = {\url{https://typesafe.ai/blog/introducing-system-one-models-and-jev}},
  note = {Official release announcement, September 15, 2026; accessed September 27, 2026}
}

@misc{palmer2026kev,
  author = {{jaredpalmer}},
  title = {{Kev-0.8B}},
  year = {2026},
  howpublished = {\href{https://huggingface.co/jaredpalmer/kev-0.8b/blob/9a45d25eb2ab761841196625383fa1dff0e56c1e/README.md}{Hugging Face model card (revision \texttt{9a45d25eb2ab})}},
  note = {Model release September 24, 2026; accessed September 27, 2026}
}

@misc{kotoba2026openjev,
  author = {{Kotoba Labs}},
  title = {{open-jev-deberta-v3-large}},
  year = {2026},
  howpublished = {\href{https://huggingface.co/com-kotobalabs/open-jev-deberta-v3-large/blob/188ee67a5c93122b916e5acd5bdb0cb3623e380a/README.md}{Hugging Face model card (revision \texttt{188ee67a5c93})}},
  note = {Accessed September 27, 2026}
}

@InProceedings{pmlr-v202-longpre23a,
  title = 	 {The Flan Collection: Designing Data and Methods for Effective Instruction Tuning},
  author =       {Longpre, Shayne and Hou, Le and Vu, Tu and Webson, Albert and Chung, Hyung Won and Tay, Yi and Zhou, Denny and Le, Quoc V and Zoph, Barret and Wei, Jason and Roberts, Adam},
  booktitle = 	 {Proceedings of the 40th International Conference on Machine Learning},
  pages = 	 {22631--22648},
  year = 	 {2023},
  editor = 	 {Krause, Andreas and Brunskill, Emma and Cho, Kyunghyun and Engelhardt, Barbara and Sabato, Sivan and Scarlett, Jonathan},
  volume = 	 {202},
  series = 	 {Proceedings of Machine Learning Research},
  month = 	 {23--29 Jul},
  publisher =    {PMLR},
  url = 	 {https://proceedings.mlr.press/v202/longpre23a.html},
}

@inproceedings{bai-etal-2025-coig,
    title = "{COIG}-{CQIA}: Quality is All You Need for {C}hinese Instruction Fine-tuning",
    author = "Bai, Yuelin  and
      Du, Xeron  and
      Liang, Yiming  and
      Jin, Leo  and
      Zhou, Junting  and
      Liu, Ziqiang  and
      Fang, Feiteng  and
      Chang, Mingshan  and
      Zheng, Tianyu  and
      Zhang, Xincheng  and
      Ma, Nuo  and
      Wang, Zekun Moore  and
      Yuan, Ruibin  and
      Wu, Haihong  and
      Lin, Hongquan  and
      Huang, Wenhao  and
      Zhang, Jiajun  and
      Lin, Chenghua  and
      Fu, Jie  and
      Yang, Min  and
      Ni, Shiwen  and
      Zhang, Ge",
    editor = "Chiruzzo, Luis  and
      Ritter, Alan  and
      Wang, Lu",
    booktitle = "Findings of the Association for Computational Linguistics: NAACL 2025",
    month = apr,
    year = "2025",
    address = "Albuquerque, New Mexico",
    publisher = "Association for Computational Linguistics",
    url = "https://aclanthology.org/2025.findings-naacl.457/",
    doi = "10.18653/v1/2025.findings-naacl.457",
    pages = "8205--8220",
    ISBN = "979-8-89176-195-7"
}

@inproceedings{yin-etal-2019-benchmarking,
    title = "Benchmarking Zero-shot Text Classification: Datasets, Evaluation and Entailment Approach",
    author = "Yin, Wenpeng  and
      Hay, Jamaal  and
      Roth, Dan",
    editor = "Inui, Kentaro  and
      Jiang, Jing  and
      Ng, Vincent  and
      Wan, Xiaojun",
    booktitle = "Proceedings of the 2019 Conference on Empirical Methods in Natural Language Processing and the 9th International Joint Conference on Natural Language Processing (EMNLP-IJCNLP)",
    month = nov,
    year = "2019",
    address = "Hong Kong, China",
    publisher = "Association for Computational Linguistics",
    url = "https://aclanthology.org/D19-1404/",
    doi = "10.18653/v1/D19-1404",
    pages = "3914--3923"
}

@misc{laurer2023universal,
  title = {{Building Efficient Universal Classifiers with Natural Language Inference}},
  author = {Laurer, Moritz and van Atteveldt, Wouter and Casas, Andreu and Welbers, Kasper},
  year = {2023},
  eprint = {2312.17543},
  archivePrefix = {arXiv},
  howpublished = {\url{https://arxiv.org/abs/2312.17543v1}},
  note = {Preprint, v1}
}

@misc{stepanov2025gliclass,
  title = {{GLiClass: Generalist Lightweight Model for Sequence Classification Tasks}},
  author = {Stepanov, Ihor and Shtopko, Mykhailo and Vodianytskyi, Dmytro and Lukashov, Oleksandr and Yavorskyi, Alexander and Yaroshenko, Mykyta},
  year = {2025},
  eprint = {2508.07662},
  archivePrefix = {arXiv},
  howpublished = {\url{https://arxiv.org/abs/2508.07662v1}},
  note = {Preprint, v1}
}

@inproceedings{zhou-etal-2025-claimgen,
  title = {{ClaimGen-CN}: A Large-scale {Chinese} Dataset for Legal Claim Generation},
  author = {Zhou, Siying and Wu, Yiquan and Chen, Hui and Hu, Xueyu and Kuang, Kun and Jatowt, Adam and Zheng, Chunyan and Wu, Fei},
  booktitle = {Findings of the Association for Computational Linguistics: EMNLP 2025},
  year = {2025},
  pages = {12296--12323},
  publisher = {Association for Computational Linguistics},
  doi = {10.18653/v1/2025.findings-emnlp.658}
}

@misc{huang2025appealcase,
  title = {{AppealCase}: A Dataset and Benchmark for Civil Case Appeal Scenarios},
  author = {Huang, Yuting and Guo, Meitong and Wu, Yiquan and Li, Ang and Liu, Xiaozhong and Yin, Keting and Sun, Changlong and Wu, Fei and Kuang, Kun},
  year = {2025},
  eprint = {2505.16514},
  archivePrefix = {arXiv},
  howpublished = {\url{https://arxiv.org/abs/2505.16514}}
}

@inproceedings{li2024lecardv2,
  title = {{LeCaRDv2}: A Large-Scale {Chinese} Legal Case Retrieval Dataset},
  author = {Li, Haitao and Shao, Yunqiu and Wu, Yueyue and Ai, Qingyao and Ma, Yixiao and Liu, Yiqun},
  booktitle = {Proceedings of the 47th International ACM SIGIR Conference on Research and Development in Information Retrieval},
  year = {2024},
  pages = {2251--2260},
  publisher = {Association for Computing Machinery},
  doi = {10.1145/3626772.3657887}
}

@inproceedings{yu2022iotmatch,
  title = {Explainable Legal Case Matching via Inverse Optimal Transport-Based Rationale Extraction},
  author = {Yu, Weijie and Sun, Zhongxiang and Xu, Jun and Dong, Zhenhua and Chen, Xu and Xu, Hongteng and Wen, Ji-Rong},
  booktitle = {Proceedings of the 45th International ACM SIGIR Conference on Research and Development in Information Retrieval},
  year = {2022},
  pages = {657--668},
  publisher = {Association for Computing Machinery},
  doi = {10.1145/3477495.3531974}
}

@article{zhao2026lcrcn,
  title = {Bridging the Gap in {Chinese} Legal Conflict Review: A Dataset, Benchmark Tasks, and Framework},
  author = {Zhao, Siwen and Xu, Yunnuo and Chen, Zhe and Qiao, Feng and Chen, Hailong and Li, XiaoRui and Lin, Sen and Ji, Zhonghang and Li, Yujun and Wang, Wei},
  journal = {Scientific Data},
  volume = {13},
  pages = {835},
  year = {2026},
  doi = {10.1038/s41597-026-07195-2}
}

@inproceedings{capp2023dataset,
  title = {{CAPP-130}: A Corpus of {Chinese} Application Privacy Policy Summarization and Interpretation},
  author = {Zhu, Pengyun and Wen, Long and Liu, Jinfei and Xue, Feng and Lou, Jian and Wang, Zhibo and Ren, Kui},
  booktitle = {Advances in Neural Information Processing Systems},
  volume = {36},
  pages = {46773--46785},
  year = {2023},
  doi = {10.52202/075280-2026}
}

@inproceedings{zhu-etal-2024-benchmarking,
  title = {Benchmarking Large Language Models on {CFLUE} - A {Chinese} Financial Language Understanding Evaluation Dataset},
  author = {Zhu, Jie and Li, Junhui and Wen, Yalong and Guo, Lifan},
  booktitle = {Findings of the Association for Computational Linguistics: ACL 2024},
  year = {2024},
  pages = {5673--5693},
  publisher = {Association for Computational Linguistics},
  doi = {10.18653/v1/2024.findings-acl.337}
}

@inproceedings{li2019chinese,
  title = {Chinese Relation Extraction with Multi-Grained Information and External Linguistic Knowledge},
  author = {Li, Ziran and Ding, Ning and Liu, Zhiyuan and Zheng, Hai-Tao and Shen, Ying},
  booktitle = {Proceedings of the 57th Annual Meeting of the Association for Computational Linguistics},
  year = {2019},
  pages = {4377--4386},
  publisher = {Association for Computational Linguistics},
  doi = {10.18653/v1/P19-1430}
}

@misc{zhu2025customer,
  title = {Evaluating, Synthesizing, and Enhancing for Customer Support Conversation},
  author = {Zhu, Jie and Dou, Huaixia and Li, Junhui and Guo, Lifan and Chen, Feng and Zhang, Chi and Kong, Fang},
  year = {2025},
  eprint = {2508.04423},
  archivePrefix = {arXiv},
  howpublished = {\url{https://arxiv.org/abs/2508.04423}}
}

@misc{xu2024fintruthqa,
  title = {{FinTruthQA}: A Benchmark Dataset for Evaluating the Quality of Financial Information Disclosure},
  author = {Xu, Ziyue and Zhou, Peilin and Shi, Xinyu and Wu, Jiageng and Jiang, Yikang and Ke, Bin and Yang, Jie},
  year = {2024},
  eprint = {2406.12009},
  archivePrefix = {arXiv},
  howpublished = {\url{https://arxiv.org/abs/2406.12009v1}},
  note = {Preprint, v1}
}

@inproceedings{sanh2022t0,
  title = {Multitask Prompted Training Enables Zero-Shot Task Generalization},
  author = {Sanh, Victor and Webson, Albert and Raffel, Colin and Bach, Stephen H. and Sutawika, Lintang and Alyafeai, Zaid and Chaffin, Antoine and Stiegler, Arnaud and Scao, Teven Le and Raja, Arun and Dey, Manan and Bari, M Saiful and Xu, Canwen and Thakker, Urmish and Sharma, Shanya Sharma and Szczechla, Eliza and Kim, Taewoon and Chhablani, Gunjan and Nayak, Nihal and Datta, Debajyoti and Chang, Jonathan and Jiang, Mike Tian-Jian and Wang, Han and Manica, Matteo and Shen, Sheng and Yong, Zheng Xin and Pandey, Harshit and Bawden, Rachel and Wang, Thomas and Neeraj, Trishala and Rozen, Jos and Sharma, Abheesht and Santilli, Andrea and Fevry, Thibault and Fries, Jason Alan and Teehan, Ryan and Bers, Tali and Biderman, Stella and Gao, Leo and Wolf, Thomas and Rush, Alexander M.},
  booktitle = {International Conference on Learning Representations},
  year = {2022},
  url = {https://openreview.net/forum?id=9Vrb9D0WI4},
  eprint = {2110.08207},
  archivePrefix = {arXiv}
}

@inproceedings{xu-etal-2020-clue,
    title = "{CLUE}: A {C}hinese Language Understanding Evaluation Benchmark",
    author = "Xu, Liang  and
      Hu, Hai  and
      Zhang, Xuanwei  and
      Li, Lu  and
      Cao, Chenjie  and
      Li, Yudong  and
      Xu, Yechen  and
      Sun, Kai  and
      Yu, Dian  and
      Yu, Cong  and
      Tian, Yin  and
      Dong, Qianqian  and
      Liu, Weitang  and
      Shi, Bo  and
      Cui, Yiming  and
      Li, Junyi  and
      Zeng, Jun  and
      Wang, Rongzhao  and
      Xie, Weijian  and
      Li, Yanting  and
      Patterson, Yina  and
      Tian, Zuoyu  and
      Zhang, Yiwen  and
      Zhou, He  and
      Liu, Shaoweihua  and
      Zhao, Zhe  and
      Zhao, Qipeng  and
      Yue, Cong  and
      Zhang, Xinrui  and
      Yang, Zhengliang  and
      Richardson, Kyle  and
      Lan, Zhenzhong",
    editor = "Scott, Donia  and
      Bel, Nuria  and
      Zong, Chengqing",
    booktitle = "Proceedings of the 28th International Conference on Computational Linguistics",
    month = dec,
    year = "2020",
    address = "Barcelona, Spain (Online)",
    publisher = "International Committee on Computational Linguistics",
    url = "https://aclanthology.org/2020.coling-main.419/",
    doi = "10.18653/v1/2020.coling-main.419",
    pages = "4762--4772"
}

@inproceedings{larson-etal-2019-evaluation,
    title = "An Evaluation Dataset for Intent Classification and Out-of-Scope Prediction",
    author = "Larson, Stefan  and
      Mahendran, Anish  and
      Peper, Joseph J.  and
      Clarke, Christopher  and
      Lee, Andrew  and
      Hill, Parker  and
      Kummerfeld, Jonathan K.  and
      Leach, Kevin  and
      Laurenzano, Michael A.  and
      Tang, Lingjia  and
      Mars, Jason",
    booktitle = "Proceedings of the 2019 Conference on Empirical Methods in Natural Language Processing and the 9th International Joint Conference on Natural Language Processing (EMNLP-IJCNLP)",
    month = nov,
    year = "2019",
    address = "Hong Kong, China",
    publisher = "Association for Computational Linguistics",
    url = "https://aclanthology.org/D19-1131/",
    doi = "10.18653/v1/D19-1131",
    pages = "1311--1316"
}

@inproceedings{fitzgerald-etal-2023-massive,
    title = "{MASSIVE}: A 1{M}-Example Multilingual Natural Language Understanding Dataset with 51 Typologically-Diverse Languages",
    author = "FitzGerald, Jack  and
      Hench, Christopher  and
      Peris, Charith  and
      Mackie, Scott  and
      Rottmann, Kay  and
      Sanchez, Ana  and
      Nash, Aaron  and
      Urbach, Liam  and
      Kakarala, Vishesh  and
      Singh, Richa  and
      Ranganath, Swetha  and
      Crist, Laurie  and
      Britan, Misha  and
      Leeuwis, Wouter  and
      Tur, Gokhan  and
      Natarajan, Prem",
    booktitle = "Proceedings of the 61st Annual Meeting of the Association for Computational Linguistics (Volume 1: Long Papers)",
    month = jul,
    year = "2023",
    address = "Toronto, Canada",
    publisher = "Association for Computational Linguistics",
    url = "https://aclanthology.org/2023.acl-long.235/",
    doi = "10.18653/v1/2023.acl-long.235",
    pages = "4277--4302"
}

@inproceedings{lai-etal-2017-race,
    title = "{RACE}: Large-scale {R}e{A}ding Comprehension Dataset From Examinations",
    author = "Lai, Guokun  and
      Xie, Qizhe  and
      Liu, Hanxiao  and
      Yang, Yiming  and
      Hovy, Eduard",
    booktitle = "Proceedings of the 2017 Conference on Empirical Methods in Natural Language Processing",
    month = sep,
    year = "2017",
    address = "Copenhagen, Denmark",
    publisher = "Association for Computational Linguistics",
    url = "https://aclanthology.org/D17-1082/",
    doi = "10.18653/v1/D17-1082",
    pages = "785--794"
}

@article{pappas-henderson-2019-gile,
    title = "{GILE}: A Generalized Input-Label Embedding for Text Classification",
    author = "Pappas, Nikolaos  and
      Henderson, James",
    journal = "Transactions of the Association for Computational Linguistics",
    volume = "7",
    year = "2019",
    address = "Cambridge, MA",
    publisher = "MIT Press",
    url = "https://aclanthology.org/Q19-1009/",
    doi = "10.1162/tacl_a_00259",
    pages = "139--155"
}

@inproceedings{wang-etal-2022-super,
    title = "Super-{N}atural{I}nstructions: Generalization via Declarative Instructions on 1600+ {NLP} Tasks",
    author = "Wang, Yizhong  and
      Mishra, Swaroop  and
      Alipoormolabashi, Pegah  and
      Kordi, Yeganeh  and
      Mirzaei, Amirreza  and
      Naik, Atharva  and
      Ashok, Arjun  and
      Dhanasekaran, Arut Selvan  and
      Arunkumar, Anjana  and
      Stap, David  and
      Pathak, Eshaan  and
      Karamanolakis, Giannis  and
      Lai, Haizhi  and
      Purohit, Ishan  and
      Mondal, Ishani  and
      Anderson, Jacob  and
      Kuznia, Kirby  and
      Doshi, Krima  and
      Pal, Kuntal Kumar  and
      Patel, Maitreya  and
      Moradshahi, Mehrad  and
      Parmar, Mihir  and
      Purohit, Mirali  and
      Varshney, Neeraj  and
      Kaza, Phani Rohitha  and
      Verma, Pulkit  and
      Puri, Ravsehaj Singh  and
      Karia, Rushang  and
      Doshi, Savan  and
      Sampat, Shailaja Keyur  and
      Mishra, Siddhartha  and
      Reddy A, Sujan  and
      Patro, Sumanta  and
      Dixit, Tanay  and
      Shen, Xudong",
    booktitle = "Proceedings of the 2022 Conference on Empirical Methods in Natural Language Processing",
    month = dec,
    year = "2022",
    address = "Abu Dhabi, United Arab Emirates",
    publisher = "Association for Computational Linguistics",
    url = "https://aclanthology.org/2022.emnlp-main.340/",
    doi = "10.18653/v1/2022.emnlp-main.340",
    pages = "5085--5109"
}

@inproceedings{li-etal-2024-cmmlu,
    title = "{CMMLU}: Measuring massive multitask language understanding in {C}hinese",
    author = "Li, Haonan  and
      Zhang, Yixuan  and
      Koto, Fajri  and
      Yang, Yifei  and
      Zhao, Hai  and
      Gong, Yeyun  and
      Duan, Nan  and
      Baldwin, Timothy",
    booktitle = "Findings of the Association for Computational Linguistics: ACL 2024",
    month = aug,
    year = "2024",
    address = "Bangkok, Thailand",
    publisher = "Association for Computational Linguistics",
    url = "https://aclanthology.org/2024.findings-acl.671/",
    doi = "10.18653/v1/2024.findings-acl.671",
    pages = "11260--11285"
}

@inproceedings{NEURIPS2023_c6ec1844,
 author = {Huang, Yuzhen and Bai, Yuzhuo and Zhu, Zhihao and Zhang, Junlei and Zhang, Jinghan and Su, Tangjun and Liu, Junteng and Lv, Chuancheng and Zhang, Yikai and Lei, Jiayi and Fu, Yao and Sun, Maosong and He, Junxian},
 booktitle = {Advances in Neural Information Processing Systems},
 doi = {10.52202/075280-2749},
 pages = {62991--63010},
 publisher = {Curran Associates, Inc.},
 title = {{C-Eval}: A Multi-Level Multi-Discipline {Chinese} Evaluation Suite for Foundation Models},
 url = {https://proceedings.neurips.cc/paper_files/paper/2023/file/c6ec1844bec96d6d32ae95ae694e23d8-Paper-Datasets_and_Benchmarks.pdf},
 volume = {36},
 year = {2023}
}

@inproceedings{khattab2020colbert,
  title = {{ColBERT}: Efficient and Effective Passage Search via Contextualized Late Interaction over {BERT}},
  author = {Khattab, Omar and Zaharia, Matei},
  booktitle = {Proceedings of the 43rd International ACM SIGIR Conference on Research and Development in Information Retrieval},
  year = {2020},
  publisher = {Association for Computing Machinery},
  pages = {39--48},
  doi = {10.1145/3397271.3401075},
  url = {https://doi.org/10.1145/3397271.3401075}
}

@inproceedings{wei2022finetuned,
  title = {Finetuned Language Models Are Zero-Shot Learners},
  author = {Wei, Jason and
      Bosma, Maarten and
      Zhao, Vincent Y. and
      Guu, Kelvin and
      Yu, Adams Wei and
      Lester, Brian and
      Du, Nan and
      Dai, Andrew M. and
      Le, Quoc V.},
  booktitle = {International Conference on Learning Representations},
  year = {2022},
  url = {https://openreview.net/forum?id=gEZrGCozdqR}
}

@misc{cai2026openjev,
  author = {Cai, Zefan},
  title = {{Open-Jev}},
  year = {2026},
  howpublished = {\url{https://github.com/Zefan-Cai/Open-Jev}},
  note = {Software release, revision \href{https://github.com/Zefan-Cai/Open-Jev/blob/3308a15ccd7eea1df7a37d6ddc39b023b801ba16/README.md}{\texttt{3308a15}}; accessed September 28, 2026}
}

@misc{lee2026semif,
  author = {Lee, Theodore},
  title = {{SemIf} (formerly {OpenJev})},
  year = {2026},
  howpublished = {\url{https://github.com/TheoLeeCJ/SemIf-OpenJev}},
  note = {Software release, revision \href{https://github.com/TheoLeeCJ/SemIf-OpenJev/blob/23cf1f39fc9534fe81437200959b6dfc7106e45a/README.md}{\texttt{23cf1f3}}; accessed September 28, 2026}
}

@misc{nimble2026,
  author = {{Bespoke Labs} and Sathiamoorthy, Maheswaran},
  title = {{Nimble}},
  year = {2026},
  howpublished = {\url{https://github.com/bespokelabsai/nimble}},
  note = {Software release, revision \href{https://github.com/bespokelabsai/nimble/blob/62076b4f2d365b5879dafcf7f6dd072a1fe76df7/README.md}{\texttt{62076b4}}; accessed September 28, 2026}
}

@misc{anyjev2026,
  author = {Zhang, Jiamu and Yang, Tianze and Shi, Yucheng and Wu, Liang},
  title = {{AnyJev}: Turn Any {LLM} into a {Jev}-Style Decision Model},
  year = {2026},
  howpublished = {\url{https://github.com/nokia-applied-research/AnyJev}},
  note = {Software release, revision \href{https://github.com/nokia-applied-research/AnyJev/blob/45add301a7aa60ed3420c83d15c061e84e5bce61/README.md}{\texttt{45add30}}; accessed September 28, 2026}
}

@misc{zhang2017encoding,
  author = {Zhang, Xiang and LeCun, Yann},
  title = {Which Encoding is the Best for Text Classification in Chinese, English, Japanese and Korean?},
  year = {2017},
  eprint = {1708.02657},
  archivePrefix = {arXiv},
  howpublished = {\url{https://arxiv.org/abs/1708.02657}}
}

@misc{utmhikari2017dmsc,
  author = {{utmhikari}},
  title = {Douban Movie Short Comments Dataset},
  year = {2017},
  howpublished = {\url{https://www.kaggle.com/datasets/utmhikari/doubanmovieshortcomments}},
  note = {Dataset, version 7; accessed September 29, 2026}
}

@article{liu2023mep,
  author = {Liu, Fan and Chen, Delong and Du, Xiaoyu and Gao, Ruizhuo and Xu, Feng},
  title = {{MEP-3M}: A Large-Scale Multi-Modal E-Commerce Product Dataset},
  journal = {Pattern Recognition},
  volume = {140},
  pages = {109519},
  year = {2023},
  doi = {10.1016/j.patcog.2023.109519},
  url = {https://doi.org/10.1016/j.patcog.2023.109519}
}

@inproceedings{zheng-etal-2019-chid,
  author = {Zheng, Chujie and Huang, Minlie and Sun, Aixin},
  title = {{ChID}: A Large-scale {Chinese IDiom} Dataset for Cloze Test},
  booktitle = {Proceedings of the 57th Annual Meeting of the Association for Computational Linguistics},
  year = {2019},
  pages = {778--787},
  publisher = {Association for Computational Linguistics},
  doi = {10.18653/v1/P19-1075},
  url = {https://aclanthology.org/P19-1075/}
}

@misc{chen2022cped,
  author = {Chen, Yirong and Fan, Weiquan and Xing, Xiaofen and Pang, Jianxin and Huang, Minlie and Han, Wenjing and Tie, Qianfeng and Xu, Xiangmin},
  title = {{CPED}: A Large-Scale Chinese Personalized and Emotional Dialogue Dataset for Conversational {AI}},
  year = {2022},
  eprint = {2205.14727},
  archivePrefix = {arXiv},
  howpublished = {\url{https://arxiv.org/abs/2205.14727}}
}

@misc{baidu2019dureaderyesno,
  author = {{Baidu}},
  title = {{DuReader Yes/No}},
  year = {2019},
  howpublished = {\url{https://github.com/baidu/DuReader}},
  note = {Opinion-polarity dataset, released December 2019; accessed September 29, 2026}
}

@inproceedings{cui-etal-2020-sentence,
  author = {Cui, Yiming and Liu, Ting and Yang, Ziqing and Chen, Zhipeng and Ma, Wentao and Che, Wanxiang and Wang, Shijin and Hu, Guoping},
  title = {A Sentence Cloze Dataset for Chinese Machine Reading Comprehension},
  booktitle = {Proceedings of the 28th International Conference on Computational Linguistics},
  year = {2020},
  pages = {6717--6723},
  publisher = {International Committee on Computational Linguistics},
  doi = {10.18653/v1/2020.coling-main.589},
  url = {https://aclanthology.org/2020.coling-main.589/}
}

@inproceedings{hu-etal-2020-ocnli,
  author = {Hu, Hai and Richardson, Kyle and Xu, Liang and Li, Lu and K{\"u}bler, Sandra and Moss, Lawrence},
  title = {{OCNLI}: Original Chinese Natural Language Inference},
  booktitle = {Findings of the Association for Computational Linguistics: EMNLP 2020},
  year = {2020},
  pages = {3512--3526},
  publisher = {Association for Computational Linguistics},
  doi = {10.18653/v1/2020.findings-emnlp.314},
  url = {https://aclanthology.org/2020.findings-emnlp.314/}
}

@inproceedings{yang-etal-2019-paws,
  author = {Yang, Yinfei and Zhang, Yuan and Tar, Chris and Baldridge, Jason},
  title = {{PAWS-X}: A Cross-lingual Adversarial Dataset for Paraphrase Identification},
  booktitle = {Proceedings of the 2019 Conference on Empirical Methods in Natural Language Processing and the 9th International Joint Conference on Natural Language Processing (EMNLP-IJCNLP)},
  year = {2019},
  pages = {3687--3692},
  publisher = {Association for Computational Linguistics},
  doi = {10.18653/v1/D19-1382},
  url = {https://aclanthology.org/D19-1382/}
}

@misc{poetry2026retrieval,
  author = {{PoetryMTEB Contributors}},
  title = {{Classical Poetry Retrieval}: Multi-Aspect Graded Retrieval for Classical Chinese Poetry},
  year = {2026},
  howpublished = {\url{https://huggingface.co/datasets/PoetryMTEB/ClassicalPoetryRetrieval}},
  note = {Dataset, version 1.3.0; accessed September 29, 2026}
}

@article{liu2023logiqa,
  author = {Liu, Hanmeng and Liu, Jian and Cui, Leyang and Teng, Zhiyang and Duan, Nan and Zhou, Ming and Zhang, Yue},
  title = {{LogiQA 2.0}---An Improved Dataset for Logical Reasoning in Natural Language Understanding},
  journal = {IEEE/ACM Transactions on Audio, Speech, and Language Processing},
  year = {2023},
  volume = {31},
  pages = {2947--2962},
  doi = {10.1109/TASLP.2023.3293046},
  url = {https://doi.org/10.1109/TASLP.2023.3293046}
}

@misc{tanchnsenticorp,
  author = {Tan, Songbo},
  title = {{ChnSentiCorp}},
  year = {n.d.},
  howpublished = {\url{https://github.com/PaddlePaddle/PaddleNLP/blob/develop/paddlenlp/datasets/chnsenticorp.py}},
  note = {Dataset distributed by PaddleNLP; accessed September 29, 2026}
}
\endgroup

\clearpage
\appendix
\section{Additional Data and Training Details}
\label{sec:appendix}
\subsection{Dataset Composition}
\label{app:data-accounting}

Table~\ref{tab:data-snapshot} reports the corpus partitions. The four benchmark partitions form Chinese-Jev Benchmark; its General component is selected from the full General test pool. Counts refer to decisions, including those derived from a shared source record. Table~\ref{tab:general-source-counts} gives the general training contributions from 20 public sources and the programmatic rule supplement.

\begin{table}[htbp]
\centering
\caption{Corpus partitions by decision type, counted in decisions. Rows labeled Benchmark make up Chinese-Jev Benchmark. The 100,000-decision General component is a subset of the full General test pool.}
\label{tab:data-snapshot}
\small
\renewcommand{\arraystretch}{1.12}
\setlength{\tabcolsep}{5pt}
\begin{tabular*}{\linewidth}{@{\extracolsep{\fill}}lrrrr@{}}
\toprule
\textbf{Partition} & \textbf{\texttt{choice}} & \textbf{\texttt{noul}} & \textbf{\texttt{score}} & \textbf{Total} \\
\midrule
\multicolumn{5}{@{}l}{\textit{General}}\\[2pt]
Train & 3,333,334 & 3,333,333 & 3,333,333 & 10,000,000 \\
Development pool & 179,723 & 215,520 & 357,543 & 752,786 \\
Calibration pool & 63,820 & 96,679 & 107,454 & 267,953 \\
Full test pool & 214,513 & 929,814 & 823,368 & 1,967,695 \\
\quad Benchmark & 33,334 & 33,333 & 33,333 & 100,000 \\
\midrule
\multicolumn{5}{@{}l}{\textit{Medical}}\\[2pt]
Train & 1,194,496 & 1,792,223 & 13,281 & 3,000,000 \\
Validation & 49,416 & 49,416 & 1,168 & 100,000 \\
Calibration & 9,846 & 9,847 & 307 & 20,000 \\
Benchmark & 98,543 & 98,544 & 2,913 & 200,000 \\
\midrule
\multicolumn{5}{@{}l}{\textit{Legal}}\\[2pt]
Train & 8,000 & 8,000 & 8,000 & 24,000 \\
Validation & 797 & 1,021 & 700 & 2,518 \\
Calibration & 317 & 450 & 250 & 1,017 \\
Benchmark & 1,500 & 1,500 & 1,300 & 4,300 \\
\midrule
\multicolumn{5}{@{}l}{\textit{Finance}}\\[2pt]
Train & 8,000 & 8,000 & 8,000 & 24,000 \\
Validation & 600 & 600 & 600 & 1,800 \\
Calibration & 200 & 200 & 200 & 600 \\
Benchmark & 1,200 & 1,200 & 1,200 & 3,600 \\
\bottomrule
\end{tabular*}
\par\vspace{4pt}
\begin{minipage}{\linewidth}
\footnotesize
Full-input limits: 1,024 tokens for General and Finance, 2,048 for Medical, and 8,192 for Legal. Medical also uses a 256-token decision-header limit.
\end{minipage}
\end{table}

\begin{table}[htb]
\centering
\caption{Source contributions to the 10 million general training decisions, combining 20 public data sources and a programmatic rule supplement. Sources are listed in descending order of contribution down each half; percentages are rounded.}
\label{tab:general-source-counts}
\small
\renewcommand{\arraystretch}{1.13}
\setlength{\tabcolsep}{4pt}
\begin{tabularx}{\linewidth}{@{}>{\raggedright\arraybackslash}Xrr@{\hspace{15pt}}>{\raggedright\arraybackslash}Xrr@{}}
\toprule
\textbf{Source} & \textbf{Decisions} & \textbf{Share} & \textbf{Source} & \textbf{Decisions} & \textbf{Share} \\
\cmidrule(r{7pt}){1-3}\cmidrule(l{7pt}){4-6}
T2Ranking~\citep{xie2023t2ranking} & 2,328,371 & 23.28\% & CMRC2019~\citep{cui-etal-2020-sentence} & 93,605 & 0.94\% \\
ASAP~\citep{bu2021asap} & 1,318,889 & 13.19\% & OCNLI~\citep{hu-etal-2020-ocnli} & 59,158 & 0.59\% \\
Dianping~\citep{zhang2017encoding} & 1,256,330 & 12.56\% & PAWS-X (Chinese)~\citep{yang-etal-2019-paws} & 45,508 & 0.46\% \\
JD full reviews~\citep{zhang2017encoding} & 973,902 & 9.74\% & Poetry retrieval~\citep{poetry2026retrieval} & 29,796 & 0.30\% \\
DMSC~\citep{utmhikari2017dmsc} & 933,902 & 9.34\% & LogiQA 2 (Chinese)~\citep{liu2023logiqa} & 11,526 & 0.12\% \\
MEP-3M~\citep{liu2023mep} & 898,261 & 8.98\% & C3~\citep{sun-etal-2020-investigating} & 11,502 & 0.12\% \\
ChID~\citep{zheng-etal-2019-chid} & 635,043 & 6.35\% & Programmatic rules & 10,700 & 0.11\% \\
CMNLI~\citep{xu-etal-2020-clue} & 587,172 & 5.87\% & MASSIVE (Chinese)~\citep{fitzgerald-etal-2023-massive} & 10,675 & 0.11\% \\
Ifeng news~\citep{zhang2017encoding} & 504,284 & 5.04\% & ChnSentiCorp~\citep{tanchnsenticorp} & 7,187 & 0.07\% \\
CPED (text)~\citep{chen2022cped} & 150,465 & 1.50\% & USTS (TED-X)~\citep{wang2023collective} & 3,007 & 0.03\% \\
DuReader Yes/No~\citep{baidu2019dureaderyesno} & 130,717 & 1.31\% &  &  &  \\
\midrule
\multicolumn{4}{@{}l}{\textbf{Total}} & \textbf{10,000,000} & \textbf{100.00\%} \\
\bottomrule
\end{tabularx}
\end{table}

\FloatBarrier

\paragraph{General benchmark component.}
We select 100,000 decisions without replacement from the General test pool, with approximately equal budgets for \texttt{choice}, \texttt{noul}, and \texttt{score}. Within each type, source tasks receive equal budgets where capacity permits; smaller tasks contribute all available examples, and unused capacity is redistributed. Selection within each task is stratified by candidate count, label, and input length. Each source contributes at most 15\% of the subset.
Table~\ref{tab:general-test-composition} groups the selected decisions using the same eight category definitions as the training corpus, listed alphabetically. All categories are covered. Their shares reflect the number of constituent source tasks and the available test data, rather than equal category quotas.

% Generated by figures/data-coverage.py from data/task-composition.json.
\begin{table}[htbp]
\centering
\caption{Task composition of the General benchmark component (100,000 decisions) and the general training corpus (10 million decisions). Categories are listed alphabetically; counts include hard and soft targets, and shares use the full size of each partition.}
\label{tab:general-test-composition}
\small
\renewcommand{\arraystretch}{1.10}
\setlength{\tabcolsep}{5pt}
\begin{tabular*}{\linewidth}{@{\extracolsep{\fill}}lrrr@{}}
\toprule
\textbf{Task category} & \textbf{Train (\%)} & \textbf{Test decisions} & \textbf{Test (\%)} \\
\midrule
Aspect mention & 4.269 & 4,480 & 4.480 \\
Aspect status and sentiment & 8.560 & 6,481 & 6.481 \\
Category and intent & 14.132 & 9,268 & 9.268 \\
Reading and reasoning & 7.624 & 11,784 & 11.784 \\
Review rating & 19.438 & 9,987 & 9.987 \\
Semantic matching & 24.067 & 26,295 & 26.295 \\
Sentiment and emotion & 14.140 & 11,961 & 11.961 \\
Textual inference and stance & 7.770 & 19,744 & 19.744 \\
\midrule
\textbf{Total} & \textbf{100.000} & \textbf{100,000} & \textbf{100.000} \\
\bottomrule
\end{tabular*}
\end{table}

\FloatBarrier

\subsection{Deduplication and Shared Material}
\label{app:deduplication}

Equation~\ref{eq:decision-fingerprint} describes exact deduplication of general decisions after task-specific conversion. Unicode NFC and whitespace normalization preserve lexical content, including numbers and negation. Equality is evaluated within a task definition, so matching text under different task identities remains separate. For \texttt{choice}, target comparison pairs each normalized candidate text with its probability before sorting. A changed answer letter caused solely by reordering does not create a label conflict. Table~\ref{tab:deduplication-cases} summarizes the resulting decisions.

\begin{table}[htb]
\centering
\caption{Exact deduplication rules for general decisions. Unless a change is specified, examples share the same task, decision type, context, and instruction. Matching inputs are merged only when their aligned targets agree; conflicting groups are discarded.}
\label{tab:deduplication-cases}
\small
\renewcommand{\arraystretch}{1.16}
\setlength{\tabcolsep}{5pt}
\begin{tabularx}{\linewidth}{@{}>{\raggedright\arraybackslash}p{.40\linewidth}>{\raggedright\arraybackslash}X@{}}
\toprule
\textbf{Variation between examples} & \textbf{Treatment} \\
\midrule
Record identifier or normalized whitespace & Same fingerprint; retain one when targets agree. \\
\texttt{choice} alternatives reordered & Same fingerprint; compare targets by candidate text, not answer letter. \\
Alternative added, removed, or replaced & Different decision, even when the question and correct answer are unchanged. \\
Same normalized input, different target & Exclude all examples with that fingerprint. \\
Different task, question, number, or negation & Different decision; preserve the changed meaning. \\
Ordered score levels changed or reordered & Different decision; the scale order is retained. \\
Shared passage, different questions & Retain distinct decisions in the same partition. \\
\bottomrule
\end{tabularx}
\end{table}

Position-dependent alternatives require special treatment. The general converter rejects detected references such as ``A and B'' or ``all of the above'' when no task-specific conversion resolves them, and rejects repeated candidate text within a \texttt{choice} question. The medical integration pipeline uses exact fingerprints that preserve candidate order, rather than the order-invariant general rule.

Split assignment operates on shared source material and its known derived examples. Distinct questions from one passage, and related views of one annotation, remain in one partition. Official held-out material takes priority over training candidates. A separate near-duplicate screen checks training candidates against indexed held-out text and excludes unresolved close matches while preserving numerical and negation differences. This screen is limited to held-out text available when a candidate is checked; paraphrases across the full training corpus are not exhaustively merged.
\FloatBarrier

\subsection{Rule-Based Synthetic Data}
\label{app:programmatic-data}

We supplement public datasets with Chinese exercises generated from templates and rules defined in this project. The general training corpus retains 10,700 such decisions (0.107\%): 10,526 concern table conditions and counts, and 174 concern rule-based grades. These counts refer to decisions, since one set of facts can support several questions.

\paragraph{Construction.}
For table tasks, we sample three to six candidate plans with prices, distances, and service availability, together with explicit selection conditions. The settings include accommodation, dining, and service packages. For grading tasks, we sample the submission status of required and optional materials for fictional registration or document-submission procedures. Chinese templates present the facts and all applicable rules. Answers are computed directly from these inputs and checked independently against the rendered text, without model-generated labels.

Some facts are marked as unknown. We retain a \texttt{score} question only when its exact count or grade is determined, and a \texttt{noul} proposition only when its truth value is determined. A \texttt{choice} question can include an insufficient-information alternative. Whether the available information determines a unique grade is itself a valid \texttt{noul} question.

\begin{figure}[htb]
\centering
\includegraphics[width=\linewidth]{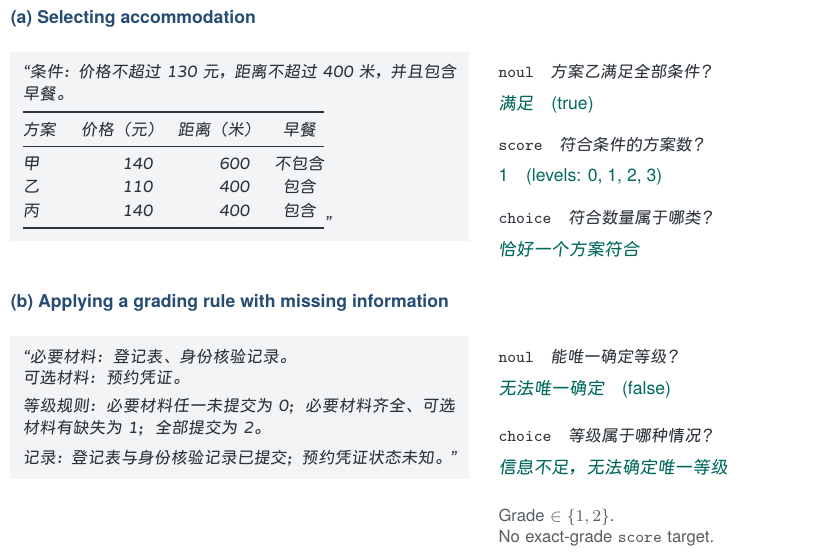}
\caption{Rule-based synthetic data. (a) Accommodation facts and selection conditions yield \texttt{choice}, \texttt{noul}, and \texttt{score} targets. (b) Missing information permits an insufficient-information \texttt{choice} target and a \texttt{noul} judgment about whether the grade is determined, but no exact-grade \texttt{score} target. Gray boxes contain the generated inputs, shortened for display; each displayed answer has target probability one. The grading rule is fictional.}
\label{fig:programmatic-examples}
\end{figure}

\paragraph{Examples.}
Figure~\ref{fig:programmatic-examples} shows two outputs of the generator. In (a), only the second accommodation plan satisfies all conditions, yielding three decisions from one table. In (b), the required materials have been submitted, but the submission status of the optional receipt is unknown. The grade could therefore be either 1 or 2: the \texttt{choice} answer is insufficient information, and the \texttt{noul} answer to whether the grade is uniquely determined is false. No exact-grade \texttt{score} question is emitted for this instance.

\subsection{Domain Data Sources}
\label{app:domain-data}

The same pipeline in Section~\ref{sec:data-pipeline} produces 3,000,000 medical training decisions and 24,000 decisions each for law and finance. Table~\ref{tab:domain-overview} reports their decision-type counts, and Table~\ref{tab:domain-sources} lists the legal and financial source contributions. The legal and financial corpora each contain 8,000 decisions per type. Medical data have a larger \texttt{noul} share because of their matching and multiple-label tasks.

\begin{table}[htbp]
\centering
\caption{Domain-specific training corpora, counted in decisions by type.}
\label{tab:domain-overview}
\small
\renewcommand{\arraystretch}{1.12}
\setlength{\tabcolsep}{5pt}
\begin{tabular*}{\linewidth}{@{\extracolsep{\fill}}lrrrr@{}}
\toprule
\textbf{Domain} & \textbf{\texttt{choice}} & \textbf{\texttt{noul}} & \textbf{\texttt{score}} & \textbf{Total} \\
\midrule
Medical & 1,194,496 & 1,792,223 & 13,281 & 3,000,000 \\
Legal & 8,000 & 8,000 & 8,000 & 24,000 \\
Finance & 8,000 & 8,000 & 8,000 & 24,000 \\
\bottomrule
\end{tabular*}
\end{table}

\begin{table}[htbp]
\centering
\caption{Legal and financial training decisions by source and decision type, after conversion and selection. Each corpus contains 8,000 decisions of each type, totaling 24,000.}
\label{tab:domain-sources}
\small
\renewcommand{\arraystretch}{1.12}
\setlength{\tabcolsep}{5pt}
\begin{tabular*}{\linewidth}{@{\extracolsep{\fill}}lrrrr@{}}
\toprule
\textbf{Source} & \textbf{\texttt{choice}} & \textbf{\texttt{noul}} & \textbf{\texttt{score}} & \textbf{Total} \\
\midrule
\multicolumn{5}{@{}l}{\textit{Legal}}\\[2pt]
AppealCase~\citep{huang2025appealcase} & 1,008 & 4,081 & 3,373 & 8,462 \\
ClaimGen-CN~\citep{zhou-etal-2025-claimgen} & 6,133 & 2,267 & 0 & 8,400 \\
LeCaRDv2~\citep{li2024lecardv2} & 200 & 200 & 2,600 & 3,000 \\
Explicit legal rules & 200 & 400 & 1,600 & 2,200 \\
LCR-CN~\citep{zhao2026lcrcn} & 302 & 898 & 0 & 1,200 \\
eCAIL~\citep{yu2022iotmatch} & 0 & 0 & 400 & 400 \\
CAPP-130~\citep{capp2023dataset} & 157 & 154 & 27 & 338 \\
\midrule
\multicolumn{5}{@{}l}{\textit{Financial}}\\[2pt]
CFLUE~\citep{zhu-etal-2024-benchmarking} & 4,100 & 3,900 & 0 & 8,000 \\
FinTruthQA~\citep{xu2024fintruthqa} & 0 & 1,000 & 5,000 & 6,000 \\
Explicit financial rules & 400 & 1,600 & 3,000 & 5,000 \\
FinRE~\citep{li2019chinese} & 2,000 & 1,000 & 0 & 3,000 \\
RoleCS~\citep{zhu2025customer} & 1,500 & 500 & 0 & 2,000 \\
\bottomrule
\end{tabular*}
\end{table}

\paragraph{Medical data.}
The medical corpus draws on 30 source families covering examinations, biomedical understanding, clinical text, and query matching. CMB~\citep{wang-etal-2024-cmb} and CMExam~\citep{NEURIPS2023_a48ad12d} supply examination questions, while CBLUE~\citep{zhang-etal-2022-cblue} contributes biomedical language-understanding tasks. Multiple-answer examinations become option-membership judgments using Equation~\ref{eq:multiselect-conversion}.

Selection checks the complete tokenized input, using limits of 2,048 tokens overall and 256 tokens for the decision header. A multiple-answer question is retained only when all of its option-wise decisions pass. Tasks with candidate sets that exceed the header budget are excluded. The resulting training set contains 3,000,000 decisions. Its 13,281 \texttt{score} examples all use the QTR query--title relevance scale.

The original medical train, validation, and test partitions remain separate. We select 100,000 validation and 20,000 calibration decisions from disjoint groups in the original validation partition. The Medical component of Chinese-Jev Benchmark contains 200,000 decisions from the original test partition.

\paragraph{Legal data.}
ClaimGen-CN provides causes of action for classification~\citep{zhou-etal-2025-claimgen}, and AppealCase provides claim-support labels and changes across appeals~\citep{huang2025appealcase}. LeCaRDv2 supplies graded case relevance~\citep{li2024lecardv2}, while eCAIL supplies grades based on legal-element overlap~\citep{yu2022iotmatch}. LCR-CN contributes judgments about conflicts between legal provisions~\citep{zhao2026lcrcn}, and CAPP-130 adds privacy-policy annotations~\citep{capp2023dataset}. The resulting tasks span classification, proposition judgments, and ordered scoring through \texttt{choice}, \texttt{noul}, and \texttt{score}.

\paragraph{Financial data.}
CFLUE supplies professional-examination questions~\citep{zhu-etal-2024-benchmarking}, and FinRE supplies entity-relation labels~\citep{li2019chinese}. RoleCS contributes customer-service strategy annotations from upstream model-generated dialogues~\citep{zhu2025customer}. FinTruthQA provides supervision for financial relevance, answer relevance, and readability~\citep{xu2024fintruthqa}. These annotations support \texttt{choice} and \texttt{noul} judgments, together with \texttt{score} targets that retain the original quality scales.

Legal and financial supplements add decisions governed by explicit fictional rules, including interest, fees, and repayment calculations in finance. Source licenses are retained, including noncommercial terms for ClaimGen-CN, AppealCase, and CFLUE.

\FloatBarrier
\subsection{RLCD Objective}
\label{app:rlcd}

The adopted RLCD objective~\citep{laya2026} evaluates perturbed candidate distributions against the same targets used by cross-entropy.

RLCD samples $G=4$ perturbations of each decision's logits. Writing one decision without the batch index, we draw $\boldsymbol{\eta}_g\sim\mathcal{N}(\mathbf{0},\sigma^2 I)$. We center each draw across candidates:
\begin{equation}
\boldsymbol{\epsilon}_g=\boldsymbol{\eta}_g-\frac{\mathbf{1}^{\top}\boldsymbol{\eta}_g}{K}\mathbf{1}.
\label{eq:centered-perturbation}
\end{equation}
Adding the centered noise to detached model logits defines the sampled logits:
\begin{equation}
\mathbf{z}_g=\operatorname{sg}(\mathbf{s})+\boldsymbol{\epsilon}_g,
\label{eq:logit-perturbation}
\end{equation}
where $\operatorname{sg}$ denotes stop-gradient. We convert each sample to a candidate distribution:
\begin{equation}
\mathbf{q}_g=\operatorname{softmax}(\mathbf{z}_g).
\label{eq:perturbed-probabilities}
\end{equation}
All perturbations reuse the same model forward pass.

The reward combines log and spherical scores to measure agreement with the target. For ordered \texttt{score} decisions, it also accounts for distance between cumulative distributions through the ranked probability score:
\begin{equation}
\operatorname{RPS}(\mathbf{q},\mathbf{y})
=\frac{1}{K-1}\sum_{j=1}^{K-1}\left(\sum_{i=1}^{j}(q_i-y_i)\right)^2.
\label{eq:ranked-probability-score}
\end{equation}
This term penalizes errors according to the ordering of levels. Combining the three terms gives the reward for each perturbed distribution:
\begin{equation}
R(\mathbf{q},\mathbf{y},t)
=\sum_{i=1}^{K}y_i\ell(q_i)
  +0.75\,\frac{\mathbf{y}^{\top}\mathbf{q}}{\lVert\mathbf{q}\rVert_2}
  -\mathbb{1}[t=\texttt{score}]\,\operatorname{RPS}(\mathbf{q},\mathbf{y}),
\label{eq:rlcd-reward}
\end{equation}
where the log score uses the floor $\ell(q_i)=\max(\log q_i,-9.21)$. The RPS term applies only to \texttt{score} decisions.

Rewards are centered within each decision's perturbation group and the resulting advantages are standardized across the batch. Denoting these detached advantages by $A_{bg}$, the policy-gradient loss is
\begin{equation}
\mathcal{L}_{\mathrm{RL}}=-\frac{1}{BG}\sum_{b=1}^{B}\sum_{g=1}^{G}
A_{bg}\left[-\frac{\lVert\mathbf{z}_{bg}-\mathbf{s}_b\rVert_2^2}{2\sigma^2}\right].
\label{eq:policy-gradient-loss}
\end{equation}
The bracketed term is the Gaussian perturbation surrogate used for the policy-gradient update. With sampled logits and advantages held fixed during differentiation, the update favors perturbations with above-average reward.

\end{document}